\documentclass[sigconf]{acmart}
\usepackage{multirow}
\usepackage{bbding}
\usepackage{color}
\usepackage{algorithm}
\usepackage{algorithmic}
\usepackage{amsmath}
\usepackage{graphics}
\usepackage{epsfig}
\usepackage{appendix}
\usepackage{enumerate}
\usepackage{tabu}
\usepackage{booktabs} 

\AtBeginDocument{%
	\providecommand\BibTeX{{%
			\normalfont B\kern-0.5em{\scshape i\kern-0.25em b}\kern-0.8em\TeX}}}

\copyrightyear{2021}
\acmYear{2021}
\setcopyright{acmcopyright}\acmConference[MM '21]{Proceedings of the 29th ACM International Conference on Multimedia}{October 20--24, 2021}{Virtual Event, China}
\acmBooktitle{Proceedings of the 29th ACM International Conference on Multimedia (MM '21), October 20--24, 2021, Virtual Event, China}
\acmPrice{15.00}
\acmDOI{10.1145/3474085.3475482}
\acmISBN{978-1-4503-8651-7/21/10}

\acmSubmissionID{1736}

\begin{document}
	\fancyhead{}
	\title{Domain Adaptive Semantic Segmentation without Source Data}
	
	
%
%
%
%

	\author{Fuming You$^{1}$, 
		Jingjing Li$^{1,2}$, 
		Lei Zhu$^{3}$, 
		Zhi Chen$^{4}$, 
		Zi Huang$^{4}$
	}\authornote{Jingjing Li is the corresponding author.}
	\affiliation{%
		\institution{$^{1}$University of Electronic Science and Technology of China, Chengdu, China}
		\city{}
		\country{}
	}
	\affiliation{%
		\institution{$^{2}$Yangtze Delta Region Institute (Huzhou), University of Electronic Science and Technology of China, Huzhou, China}
		\city{}
		\country{}}
	\affiliation{%
		\institution{$^{3}$Shandong Normal University, Jinan, China}
		\city{}
		\country{}
	}
	\affiliation{%
		\institution{$^{4}$University of Queensland, Brisbane, Australia}
		\city{}
		\country{}
	}
	\email{fumyou13@gmail.com, 
		lijin117@yeah.net, 
		leizhu0608@gmail.com, 
		zhi.chen@uq.edu.au, 
		huang@itee.uq.edu.au}

	%
	%
	%
	%

	
	\begin{abstract}
		Domain adaptive semantic segmentation is recognized as a promising technique to alleviate the domain shift between the labeled source domain and the unlabeled target domain in many real-world applications, such as automatic pilot. However, large amounts of source domain data often introduce significant costs in storage and training, and sometimes the source data is inaccessible due to privacy policies.
		To address these problems, we investigate domain adaptive semantic segmentation without source data, which assumes that the model is pre-trained on the source domain, and then adapting to the target domain without accessing source data anymore. Since there is no supervision from the source domain data, many self-training methods tend to fall into the ``winner-takes-all'' dilemma, where the {\it majority} classes totally dominate the segmentation networks and the networks fail to classify the {\it minority} classes. Consequently, we propose an effective framework for this challenging problem with two components: positive learning and negative learning. 
		In positive learning, we select the class-balanced pseudo-labeled pixels with intra-class threshold, while in negative learning, for each pixel, we investigate which category the pixel does not belong to with the proposed heuristic complementary label selection.
		Notably, our framework can be easily implemented and incorporated with other methods to further enhance the performance. Extensive experiments on two widely-used synthetic-to-real benchmarks demonstrate our claims and the effectiveness of our framework, which outperforms the baseline with a large margin.
		Code is available at \url{https://github.com/fumyou13/LDBE}.
	\end{abstract}

	
	
	
	\maketitle
	\section{Introduction}
	In recent years, deep convolutional neural networks have achieved significant success across various multimedia tasks, such as cross-modal retrieval~\cite{wang2017adversarial}, image captioning~\cite{guo2019aligning} and so on. However, the impressive success heavily relies on the abundant labeled training data and the model fails to generalize to the novel and unseen instances. As a practical alternative, unsupervised domain adaptation (UDA) enables transferring knowledge from a labeled source domain to an unlabeled target domain, where different domains have distinctive distributions.
	
	Semantic segmentation is a challenging task in real-world applications, which aims at assigning a semantic label to each pixel in an image. In practice, deep convolutional neural networks have achieved exciting performance on semantic segmentation, but heavily rely on the sufficient manual annotations, which are more expensive and time-consuming compared to other applications, e.g., object recognition. To handle this, many researchers turn to the synthetic datasets since the ground-truth labels are available, and transfer the knowledge learned from synthetic datasets to real-world applications with the help of UDA~\cite{ben2010theory,li2019cycle,li2019transfer}. This paradigm is called \emph{Domain Adaptive Semantic Segmentation}~\cite{hoffman2017cycada,luo2019taking} (DASS).
	
	Domain adaptive semantic segmentation has got great attention and various methods have been proposed, which can be roughly divided into two categories: adversarial training~\cite{chen2019crdoco,chang2019all,tsai2018learning,vu2019advent,saito2018maximum} and self-training~\cite{zhang2019category,zou2018unsupervised,zou2019confidence,li2020content}. Adversarial training based methods usually learn domain-invariant feature representations to achieve adaptation, while self-training based methods usually mitigate the domain gap iteratively through various strategies. Notably, many works~\cite{pan2020unsupervised,wang2020differential,yang2020fda,yu2021dast} integrate both of them to achieve better performance.
	
\begin{figure}[t]
		\centering
		\includegraphics[width=\linewidth]{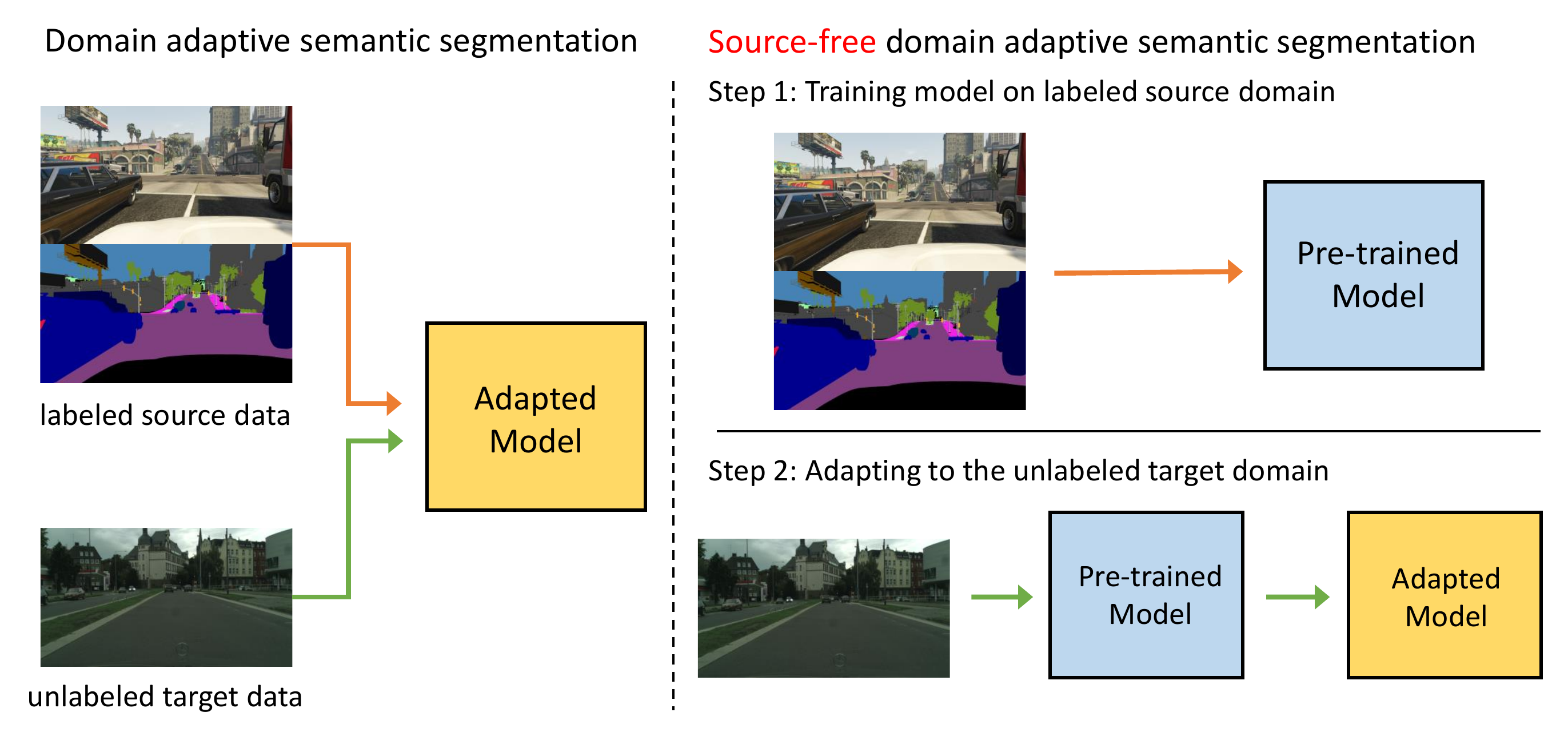}
		\vskip -7pt
		\caption{\small (Best viewed in color.) The comparison between domain adaptive semantic segmentation (DASS) and source-free domain adaptive semantic segmentation (SFSS). In DASS, both source and target domain data should be accessible during training. However, in SFSS, the training procedure is divided into two steps: (1) Training model on labeled source domain. (2) Adapting to the unlabeled target domain. Compared to conventional DASS, SFSS shows superiority on privacy protection and data transmission.}
		\label{fig:compare1}
		\vskip -10pt
	\end{figure}

	However, it is worth noting that DASS is still suffering several limitations: (1) When adapting to the target domain, accessing the large amounts of source data is inefficient and impractical. For instance, The size of the synthetic dataset GTA5 is nearly 61.6GB, which is usually quite hard to store and transmit. (2) The privacy of source data cannot be guaranteed, especially in the sensitive scenarios, e.g., medical image segmentation~\cite{mahmood2018unsupervised}. Therefore, it is necessary to investigate source-free domain adaptation (SFDA) in semantic segmentation. Specifically, we divide the training process into two steps: training the supervised model on the labeled source domain, and then adapting to the unlabeled target domain based on the pre-trained model. For the sake of building an efficient and privacy-preserving domain adaptive semantic segmentation model~\cite{li2021faster}, we propose a practical scenario named \emph{Source-Free domain adaptive Semantic Segmentation} (SFSS). The comparison between DASS and SFSS is illustrated in Fig.~\ref{fig:compare1}.

	
	In SFSS, accessing the source data is prohibited during adapting to the target domain, i.e., the domain divergence~\cite{ben2010theory,li2020maximum} between source and target domains is unknown~\cite{liang2020we}. Thus, most DASS methods focusing on mitigating the domain divergence to achieve adaptation~\cite{luo2019taking,tsai2018learning} would fail to generalize to SFSS. And recent SFDA methods are also not suitable for SFSS. Specifically, recent SFDA methods can be divided into two categories: GAN based~\cite{li2020model} and pseudo-label based~\cite{liang2020we,kim2020domain,li2020free}. However, in semantic segmentation, (1) generating a pixel-level image is impractical, (2) and instance-level pseudo-label methods may fail in the dense task, whose objective is a pixel rather than an image. We will elaborate the comparison with other methods in Sec.~\ref{sec:methods} and Sec.~\ref{sec:result}. Furthermore, in the dense task semantic segmentation, pseudo-label based model tends to suffering from the ``winner-takes-all'', which indicates that the model tends to over-fitting the {\it majority} classes, but ignoring the {\it minority} classes. This phenomenon becomes more serious in SFSS, while DASS allows access to the labeled data in source domain during adaptation, thus preventing the {\it majority} classes totally dominate the model and preserving the {\it minority} classes. More details are elaborated in Sec.~\ref{sec:pos} and Fig.~\ref{fig:fig2}.
	
	To address above issues, we propose an effective framework for SFSS, which contains two mutually reinforcing components: positive learning and negative learning. In positive learning, we analyze in detail why the conventional self-training methods perform poorly in this setting and highlight the challenges of SFSS, then we introduce an intra-class threshold to filter the class-balanced pseudo-labeled pixels to prevent the ``winner-takes-all''. In negative learning, we propose a heuristic complementary label selection (HCLS) to generate the complementary label for each pixel. The complementary label indicates which category the pixel does not belong to.
	Intuitively, applying the source pre-trained model to the target domain directly leads to lots of noisy-labeled pixels due to the domain gap. Therefore, learning from the complementary label is a more proper strategy since there exists less noises in the complementary label. Extensive experiments demonstrate that each component is effective in improving the performance, and that combining both of them leads to better results. In a nutshell, we have following contributions:
	\begin{enumerate}[1)]
		\item We propose a novel and challenging scenario: \emph{source-free domain adaptive semantic segmentation}, which provides a path of training an efficient and privacy-preserving semantic segmentation model.
		To the best of our knowledge, we are at the very early attempt to investigate the source-free domain adaptation in semantic segmentation.
		\item We highlight the main challenges of SFSS, and point out why many mainstream self-training and source-free domain adaptation methods fail in this setting. Then, We re-implement them under SFSS setting, and the empirical results demonstrate our claims. Based on the observation of challenges, we propose an effective label-denoising framework named LD, which addresses this problem with two mutual-reinforcing components: positive learning and negative learning. The framework can be easily implemented and incorporated with other methods (e.g., attention module) to further enhance the performance.
		\item We conduct experiments on two synthetic-to-real benchmarks: GTA5$\rightarrow$Cityscapes and SYNTHIA$\rightarrow$Cityscapes. Our method outperforms all comparisons with a significant improvement. Particularly, LD outperforms the baseline over $9.8\%$ mIoU in GTA5$\rightarrow$Cityscapes and $12.5\%$ mIoU in SYNTHIA$\rightarrow$Cityscapes, respectively. Furthermore, our method achieves competitive performance compared to the source-accessible methods, and shows superiority on privacy-preserving, data transmission and efficiency.
	\end{enumerate}
	
	\section{Related Work}
	\subsection{Source-free Domain Adaptation}
	Unsupervised domain adaptation (UDA) has witnessed significant development in both theory and various applications recently~\cite{ben2010theory,zhao2019learning,luo2019taking,li2019locality,li2018heterogeneous}. Domain adaptation aims at learning a general model from both labeled source domain and unlabeled target domain, where different domain has distinctive distributions~\cite{long2015learning,sun2016deep}. 
	
	By considering the privacy and data transmission, source-free domain adaptation (SFDA) attracts a lot of attention lately~\cite{liang2020we,kim2020domain}. SFDA divides the training process into two steps: training the model on labeled source domain, and then adapt to the unlabeled target domain. Although SFDA is proposed very recently, it has been studied in many fields. For object recognition, 3C-GAN~\cite{li2020model} generated target-style samples and leverages other regularizations to moderately retrain the source model. Hou et al.~\cite{hou2020source} converted the target-style images to the source-style based on the mean and variance stored in BN layers of the source model. SFDA~\cite{kim2020domain} proposed to use the samples with low entropy to refine the pseudo labels for training. SHOT~\cite{liang2020we} proposed two strategies: information maximum loss and weighed deep clustering, and achieved state-of-the-art performance. For object detection, SFOD~\cite{li2020free} proposed a metric named self-entropy decent to select the appropriate threshold for pseudo label generation. 
	
	To the best of our knowledge, we are at the very early attempt to investigate SFDA in semantic segmentation. In Sec.~\ref{sec:result}, we compare our method with other state-of-the-art SFDA methods in detail.
	\subsection{Domain Adaptive Semantic Segmentation}
	Domain Adaptive semantic segmentation (DASS) is an important application of UDA. Existing DASS methods can be roughly divided into two categories: adversarial training and self-training. 
	
	Adversarial training mainly includes two strategies: (1) Employing a style transfer across domains through adversarial training~\cite{chang2019all,zhang2018fully}. (2) Learning the domain-invariant feature representations across domains~\cite{tsai2018learning,vu2019advent,tsai2019domain,luo2019taking,saito2018maximum}. For example, Vu et al.~\cite{vu2019advent} match the entropy of output predictions in source and target domains via adversarial training. Recently, Luo et al.~\cite{luo2021category} proposed two modules to purify the features and perform the category-wise adversarial training based on it. However, these methods cannot be implemented directly in SFSS setting since the source data are unknown during training. Furthermore, adversarial based methods usually take more training time, while our method provides an efficient cross-domain semantic segmentation model (see Sec.~\ref{sec:eff}). 
	
	Another mainstream line of work for DASS leverages the self-training. Some methods~\cite{chen2019domain,zou2019confidence,vu2019advent} utilizes prediction entropy to guarantee a distinct decision boundary in target domain. Recently, many methods product pseudo-labels in target domain based on confidence or uncertainty estimation~\cite{iqbal2020mlsl,lian2019constructing,pan2020unsupervised,sakaridis2019guided,zheng2021rectifying,zou2018unsupervised,li2020content,zhang2019category}. For example, Zou et al.~\cite{zou2018unsupervised} proposed a class-balanced selection strategy and selected the high-confidence pseudo-labeled target samples, while Zheng et al.~\cite{zheng2021rectifying} estimated the uncertainty via an auxiliary classifier and selected the rectified pseudo labels through the dynamic threshold. Unlike the aforementioned methods, Li et al.~\cite{li2020content} proposed to select the source images that share similar distribution with the real ones in the target domain to alleviate the domain gap in another perspective. 
	
	The most related work is CBST~\cite{zou2018unsupervised}, which proposed class-balanced pseudo-labeling. In our proposed label-denoising, intra-class threshold selection is also adopted. However, we have significant different motivations and claimed contributions. In SFSS, without any supervision from source domain, the class-imbalance issue becomes more severe than DASS. Once the category is a majority, it will be gradually replaced and totally dominated if there is no external intervention. In DASS, the supervision from source domain data preserve the minority classes, thus alleviating this issue. We summarize this phenomenon as “winner-takes-all”. Therefore, based on our original findings on the challenges of SFSS, class-balanced selection is a simple yet significant way to address it and it is not a direct re-use. Besides, we also propose negative learning to address other challenges.
	
	
	\begin{figure*}
		\centering
		\includegraphics[width=0.93\linewidth]{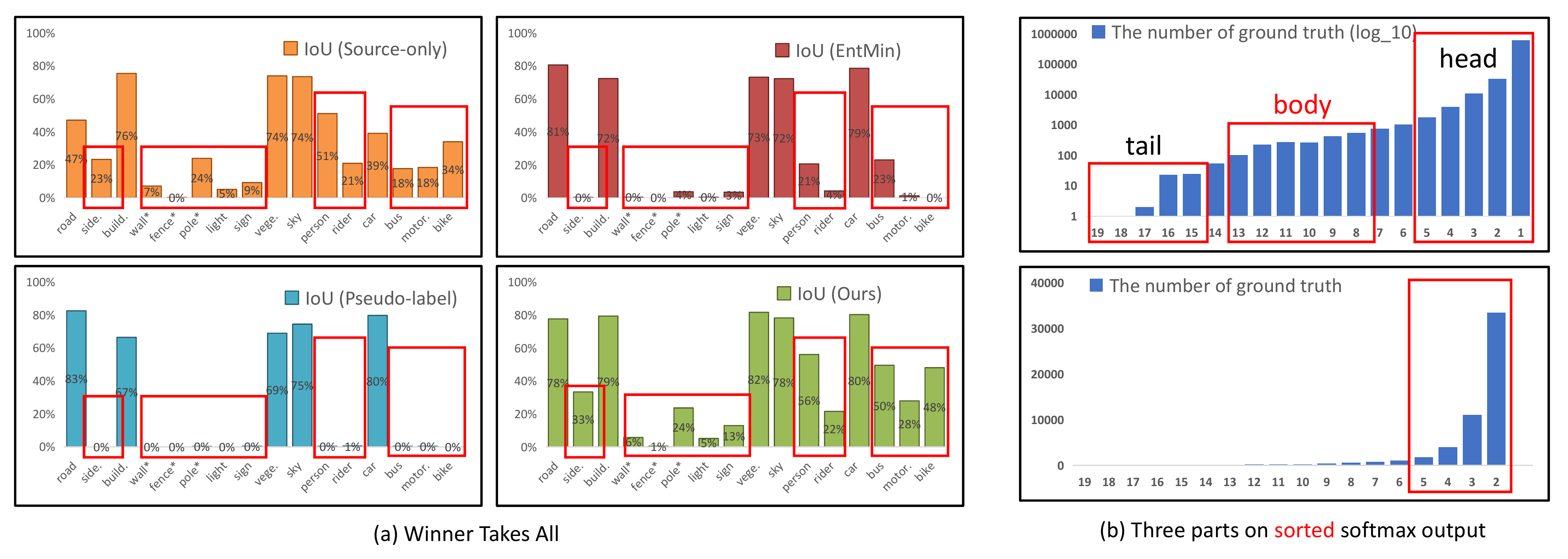}
		\vskip -10pt
		\caption{\small (Best viewed in color.) (a) The illustration of the results of different approaches on SYNTHIA$\rightarrow$Cityscapes. Conventional approaches suffer from the ``winner-takes-all'', while our method consistently improves the performance on each classes. (b) We randomly select a batch (two 600 $\times$ 600 images in a batch) in GTA5$\rightarrow$Cityscapes for visualization. In figure~(b), $(x,y)$ indicates that there are $y$ samples' true classes correspond to the Top $x$ softmax outputs. Notably, we adopt the logarithmic coordinate system with a base of 10 in the top figure, while in the bottom figure, we remove the Top 1 softmax output in another figure for better presentation since it is an order of magnitude above the rest.}
		\label{fig:fig2}
	\end{figure*}
	\section{Approach}
	In this section, we present the challenges of SFSS, and show why existing methods fail in this setting. Then, based on the observations, we address it with two components: positive learning and negative learning. Notably, our framework only contains the loss function of label denoising, which adds few burdens to the complexity of the model and the running time. Furthermore, our framework can be easily implemented and incorporated with other methods (e.g., attention module) to achieve better performance.
	\subsection{Notations and Definitions}
	In domain adaptive semantic segmentation, we have a source domain $\mathcal{X}_{S}\subset \mathbb{R}^{H\times W\times 3}$ with corresponding ground-truth segmentation maps $\mathcal{Y}_{S}\subset \{(0,1)\}^{H\times W\times C}$, where $H, W, C$ denote the height, width and the number of classes, respectively. The target domain $\mathcal{X}_{T}\subset \mathbb{R}^{H\times W\times 3}$ is unlabeled. Let $S$ be a segmentation network with parameters $\theta$ , which takes an image $x$ and gives a prediction $p^{(h,w,c)}$. Especially, in SFSS, the training procedure has two stages: training a supervised loss on the labeled source domain and adapting to the unlabeled target domain. For simplicity, we optimize the cross-entropy loss over the source domain:
	\begin{equation}
	\label{eq:ce}
	\mathcal{L}_{ce}=-\sum_{(x,y)\in(\mathcal{X}_S,\mathcal{Y}_S)}y\log S(x,\theta).
	\end{equation}
	However, directly applying the model trained on source domain to the target domain often results in significant performance degradation~\cite{ben2010theory,li2020content}. Therefore, we present our method and show how to enhance the performance on target domain without accessing source data.
	\subsection{Positive Learning}
	\label{sec:pos}
	Pseudo labeling is a widely used strategy in semi-supervised learning, and attracts great attention in DASS recently. Intuitively, obtaining the pseudo labels through applying argmax operation on the softmax predictions is unreliable due to the domain gap between source and target domains. A trivial solution is to select the high-confidence pseudo labels, which effectively removes the wrong labels but neglects the ``winner-takes-all'' dilemma, where the model would be bias towards the {\it majority} classes and ignore the {\it minority} classes (see Fig.~\ref{fig:fig2}~(a)). Therefore, we propose to select the pixels with intra-class confidence. To avoid the imbalanced selection, the intra-class threshold is defined as:
	\begin{equation}
	\delta^{(c)} =  \top_\alpha(\mathcal{P}_{T}^{(c)}),
	\label{eq:alpha}
	\end{equation}
	where $\top_\alpha$ means the top $\alpha$ (\%) value in the softmax prediction value set $\mathcal{P}_{T}^{(C)}$ with respect to class $c\in C$. Then, for each class $c\in C$, we select the samples whose softmax confidence is larger than $\delta^{(c)}$. It is worth noting that our selection strategy addresses the aforementioned two challenges: (1) The most noise labels are filtered due to the high softmax prediction confidence selection. (2) The intra-class confidence selection addresses the dilemma where selecting high-confidence pseudo labels always results in the biased segmentation network. We update the pseudo labels at the beginning of each epoch to avoid additional running time. After selection, we optimize following cross-entropy loss at the selected pixels:
	\begin{equation}
	\mathcal{L}_{sce} = -\sum_{(x,\hat{y})\in(\mathcal{X}'_{T},\mathcal{\hat{Y}}'_{T})} \hat{y}\log S(x,\theta),
	\end{equation}
	where $(\mathcal{X}'_{T},\mathcal{\hat{Y}}'_{T})$ are the selected pixels on target domain and the pseudo labels are obtained by the argmax operation based on softmax prediction. Furthermore, we optimize the pixels not be selected by entropy minimization. Entropy minimization has been shown to be effective in semi-supervised learning~\cite{grandvalet2005semi,springenberg2015unsupervised} and domain adaptation~\cite{vu2019advent,you2021learning}. Vu et al.~\cite{vu2019advent} argues that entropy minimization can be recognized as a soft-assignment version of cross-entropy loss. Therefore, to balance the unselected pixels, we optimize the following entropy minimization loss: 
	\begin{equation}
	\mathcal{L}_{ent} = -\sum_{p\in \mathcal{P}_{T}}\sum_{c=1}^{C} p^{(c)} \log p^{(c)},
	\end{equation}
	where $\mathcal{P}_{T}$ is the set of softmax prediction over each pixels on target domain images. Now, we summarize the formulation from positive learning perspective:
	\begin{equation}
	\label{eq:pos}
	\mathcal{L}_{pos} = \mathcal{L}_{sce}+\lambda_{ent}\mathcal{L}_{ent}
	\end{equation} 
	\begin{algorithm}[tb] 
		\caption{Pseudo code for HCLS} 
		\label{alg:hcls} 
		\begin{algorithmic}[1] 
			\REQUIRE 
			The sorted softmax output on target domain $p_T\in \mathcal{P}_T$, the number of classes $C$, the width of selection range $\epsilon$.
			\ENSURE 
			The complementary label $\overline{y}$; 
			
			\STATE $K= \lfloor C/2 \rfloor$ + rand ($-\epsilon,+\epsilon$)
			\label{eq:epsilon}
			\STATE $\overline{y}$= The class having the Top $K$ softmax value.
			\RETURN $\overline{y}$; 
		\end{algorithmic} 
		
	\end{algorithm}
	\vspace{-4mm}
	\subsection{Negative Learning}
	\label{sec:neg}
	In order to avoid the incorrect pseudo labels, we propose to address SFSS from another perspective, i.e., negative learning. Our idea is easy to be understood: instead of determining which category a pixel belongs to, it is easier to infer which category it does not belong to, which is more feasible in SFSS where no supervised information is provided. Let $x\in\mathcal{X}_{T}$ be an input, and $\hat{y}\in\mathcal{\hat{Y}}_T$ be its corresponding pseudo label. We generate the complementary label $\overline{y}\in\mathcal{\overline{Y}}_{T}$($\overline{y} \neq \hat{y}$), and optimize following loss function:
	\begin{equation}
	\label{eq:neg}
	\mathcal{L}_{neg} = -\sum_{(x,\overline{y})\in(\mathcal{X}_{T},\mathcal{\overline{Y}}_{T})} \overline{y}\log(1-S(x,\theta))
	\end{equation}
	To the best of our knowledge, we are the first to perform negative learning in pixel-level. In previous work, Kim et al~\cite{kim2019nlnl} perform negative learning in classification task and the complementary label is generated randomly from the labels of all classes except for the pseudo label $\hat{y}$. However, we further raise following question: which soft-label deserves to be optimized? To answer this question, we illustrate a toy example in Fig.~\ref{fig:fig2}~(b). Firstly, if we sort the softmax output $\mathcal{P}_T$, the class having the max softmax value is selected as the pseudo label. However, in SFSS, there are lots of noises in the pseudo labels due to the domain gap. Therefore, intuitively, the ``head'' classes (except the pseudo label) have a high probability of being the correct label and performing negative learning on these complementary labels causes an accumulation of errors. And the loss on the complementary labels inside the ``tail'' classes converges to 0. Therefore, we propose a heuristic complementary label selection (HCLS) to select the more accurate and valuable complementary labels, i.e., the ``body'' classes, which avoids the risks mentioned above. The pseudo code of HCLS in presented in Algorithm \ref{alg:hcls}.
		\begin{table*}[t]\vspace{-0.1in}
			\caption{\small Results of GTA5$\rightarrow$Cityscapes.'$\dag$' means the results are based on our implementation. The mechanism 'S', 'A' and 'T' stand for self-training, adversarial training and image-to-image translation, respectively. 'SF' means whether the method is evaluated under source-free setting. ``gain'' indicates the mIoU improvement over using the source only under source-free settings. The best results under source-free settings are highlighted by bold and the best results under both source-free and source-accessible settings are highlighted by underlining.}
			
			\label{table:gta5}
			\vspace{-3.5mm}
			\begin{center}
				\begin{small}
					\resizebox{\textwidth}{32mm}{
						\begin{tabular}{l|c|c|ccccccccccccccccccc|cc}
							\toprule[1.5pt]
							\multicolumn{24}{c}{GTA5$\rightarrow$Cityscapes}\\
							\toprule
							Method &\rotatebox{90}{mech.}&SF& \rotatebox{90}{road}&\rotatebox{90}{side.}&\rotatebox{90}{build.}&\rotatebox{90}{wall*}&\rotatebox{90}{fence*}&\rotatebox{90}{pole*}&\rotatebox{90}{light}&\rotatebox{90}{sign}&\rotatebox{90}{vege.}&\rotatebox{90}{terr.}&\rotatebox{90}{sky}&\rotatebox{90}{pers.}&\rotatebox{90}{rider}&\rotatebox{90}{car}&\rotatebox{90}{truck}&\rotatebox{90}{bus}&\rotatebox{90}{train}&\rotatebox{90}{motor}&\rotatebox{90}{bike}&\rotatebox{90}{\bf mIoU}&\rotatebox{90}{\bf gain}\\
							\midrule
							Source only\dag&-&\multirow{9}{*}{\Checkmark}&60.6&17.4&73.9&17.6&20.6&21.9&31.7&15.3&79.8&18.1&71.1&55.2&22.8&68.1&32.3&13.8&3.4&\bf\underline{34.1}&21.2&35.7&-\\
							EntMin(CVPR'19)\dag&S&&82.8&0.0&70.2&2.2&0.3&0.4&2.8&1.6&79.9&8.1&79.2&22.2&0.1&83.1&22.5&30.0&2.0&6.3&0.0&26.0&\textcolor{red}{-9.7 \%}\\
							Pseudo\dag&S&&83.2&0.0&67.3&1.1&0.0&0.1&1.2&1.2&77.7&1.3&81.4&11.5&0.0&81.7&18.0&14.8&0.0&3.7&0.0&23.4&\textcolor{red}{-12.3\%}\\
							Pse.+Ent.\dag&S&&83.0&0.0&66.0&0.2&0.0&0.0&0.4&0.5&75.4&0.0&82.3&7.3&0.0&80.8&12.5&2.6&0.0&2.2&0.0&21.8&\textcolor{red}{-13.9\%}\\
							Pse.+Sel.\dag&S&&83.2&0.8&76.1&13.5&7.9&4.4&9.2&5.9&82.9&27.3&77.2&41.0&1.8&\bf83.8&36.3&45.8&\bf5.0&15.8&0.0&32.5&\textcolor{red}{-3.2\%}\\
							SHOT(ICML'20)~\cite{liang2020we}\dag&S&&87.6&44.4&80.6&24.4&19.4&9.8&14.4&9.6&83.5&37.6&79.8&49.6&0.0&78.6&36.7&50.1&\bf8.0&18.0&0.0&38.5&\textcolor{green}{+2.8\%}\\
							LD w/o $\mathcal{L}_{neg}$(ours)\dag &S&&89.6&46.1&66.6&30.7&8.7&\bf28.7&\bf32.8&\bf29.7&81.5&36.5&\bf83.4&57.0&26.7&82.9&28.9&31.5&0.3&16.5&\bf38.3&43.0&\textcolor{green}{+7.3\%}\\
							LD w/o $\mathcal{L}_{pos}$(ours)\dag&S&&83.9&29.8&79.5&27.9&\bf21.3&23.6&25.9&19.5&79.2&27.3&71.7&\bf58.0&\bf28.2&82.3&29.2&44.9&\bf5.0&29.2&18.6&41.3&\textcolor{green}{+5.6 \%}\\
							LD(ours)\dag&S&&\bf91.6&\bf53.2&\bf80.6&\bf36.6&14.2&26.4&31.6&22.7&\bf83.1&\underline{\bf42.1}&79.3&57.3&26.6&82.1&\bf41.0&\bf50.1&0.3&25.9&19.5&\bf45.5&\textcolor{green}{+9.8 \%}\\
							
							\midrule
							SIBAN(ICCV'19)~\cite{luo2019significance}& A & \multirow{8}{*}{\XSolidBrush}&  88.5& 35.4& 79.5& 26.3& 24.3& 28.5& 32.5& 18.3& 81.2& 40.0& 76.5& 58.1& 25.8& 82.6& 30.3& 34.4& 3.4& 21.6& 21.5& 42.6&-\\
							AdaptSeg(CVPR'18)~\cite{tsai2018learning}&A& &87.3& 29.8& 78.6& 21.1& 18.2& 22.5& 21.5& 11.0& 79.7& 29.6& 71.3& 46.8& 6.5& 80.1& 23.0& 26.9& 0.0& 10.6& 0.3& 35.0&-\\
							CLAN(PAMI'21)~\cite{luo2021category}& A& &88.7& 35.5 &80.3& 27.5& 25.0& 29.3& 36.4 &28.1 &84.5 &37.0& 76.6& 58.4 &\underline{29.7}& 81.2 &38.8& 40.9& 5.6 &32.9 &28.8 &45.5&-\\
							DPR(ICCV'19)~\cite{tsai2019domain} & SAT& & 92.3& 51.9& 82.1& 29.2& 25.1 &24.5& 33.8& \underline{33.0}& 82.4 &32.8& 82.2 &58.6& 27.2& 84.3& 33.4& 46.3& 2.2& 29.5& 32.3& 46.5&-\\
							
							IntraDA(CVPR'20)~\cite{pan2020unsupervised}& SA& & 90.6& 37.1& 82.6& 30.1& 19.1& 29.5& 32.4& 20.6& \underline{85.7}& 40.5& 79.7& 58.7& 31.1& 86.3& 31.5& 48.3& 0.0& 30.2& 35.8& 46.3&-\\
							CRST(ICCV'19)~\cite{zou2019confidence}&S& & 91.0& 55.4& 80.0 &33.7 &21.4& 37.3 &32.9& 24.5& 85.0& 34.1& 80.8 &57.7& 24.6& 84.1& 27.8& 30.1 &\underline{26.9}& 26.0& 42.3& 47.1&-\\
							DAST(AAAI'21)~\cite{yu2021dast}&SA&&92.2 &49.0& 84.3& 36.5 &\underline{28.9}& \underline{33.9} &\underline{38.8}& 28.4& 84.9 &41.6& 83.2& \underline{60.0} &28.7& \underline{87.2}& 45.0& 45.3& 7.4 &33.8& 32.8& 49.6&-\\
							CCM(ECCV'20)~\cite{li2020content}&S&&\underline{93.5} &\underline{57.6} &\underline{84.6}& \underline{39.3}& 24.1& 25.2& 35.0& 17.3& 85.0& 40.6 &\underline{86.5}& 58.7& 28.7& 85.8& \underline{49.0} &\underline{56.4}& 5.4 &31.9& \underline{43.2} &\underline{49.9}&-\\
							
							\bottomrule[1.5pt]
						\end{tabular}}
						\vskip -7pt
					\end{small}
				\end{center}
			\end{table*}
	\subsection{Overall Formulation}
	Actually, performing positive learning or negative learning individually improves the performance of the source pre-trained model on target domain (see Table~\ref{table:gta5},~\ref*{table:synthia}). Furthermore, incorporating them achieves better performance. On one hand, positive learning provides the class-balanced pseudo labels and helps correcting the complementary labels, thus improving the negative learning. On the other hand, negative learning denoises the pseudo labels, preventing the accumulation of errors in pseudo-labeling. Therefore, these two components are not mutually exclusive or independent, but reinforcing. The overall objective function for segmentation networks $S$ becomes:
	\begin{align}
	\label{eq:overall}
	\mathcal{L}_{LD}&=\mathcal{L}_{pos}+\lambda_{neg}\mathcal{L}_{neg}\\ \notag
	&=\mathcal{L}_{sce}+\lambda_{ent}\mathcal{L}_{ent}+\lambda_{neg}\mathcal{L}_{neg}
	\end{align}
	where $\lambda_{ent}$ and $\lambda_{neg}$ are the hyper-parameters. And it is worth noting that the $\mathcal{L}_{sce}$ is calculated on the selected pixels, and the selected pixels are updated at the beginning of each epoch out of saving training time. And $\mathcal{L}_{ent}$ and $\mathcal{L}_{neg}$ are calculated on all pixels since both of them not explicitly provide supervision  and can balance the unselected pixels. The details of training process are described in Algorithm~\ref{alg:pn}.
	\begin{algorithm}[t] 
		\caption{Pseudo code for the proposed LD} 
		\label{alg:pn} 
		\begin{algorithmic}[1] 
			\REQUIRE The labeled source domain \{$\mathcal{X}_{S},\mathcal{Y}_{S}$\}, the unlabeled target domain \{$\mathcal{X}_{T}$\}, the selection range $\alpha$ and $\epsilon$, the segmentation networks $S$ and the hyper-parameters $\lambda_{ent}$ and $\lambda_{neg}$.
			
			\STATE Training the segmentation networks $S$ on the labeled source domain with loss function Eq.~(\ref{eq:ce}).
			\FOR{ iteration=1 to max\_iteration}
			\STATE Input $x_{T} \in \mathcal{X}_{T}$ to S, obtain the softmax predictions.
			\STATE Calculate the class-wise threshold according to Eq.~(\ref{eq:alpha}), and select the class-balanced pseudo labels.
			\STATE Calculate the loss function of positive learning according to Eq.~(\ref{eq:pos}).
			\STATE Generate the complementary labels according to Algorithm $\ref{alg:hcls}$.
			\STATE Calculate the loss function of negative learning according to Eq.~(\ref{eq:neg}).
			\STATE Update the parameters of segmentation networks $\theta$ according to Eq.~(\ref{eq:overall})
			\ENDFOR 
			\RETURN $S$. 
		\end{algorithmic} 
	\end{algorithm}
	
	\section{Experiments}

		We evaluate the proposed method on two synthetic-to-real tasks: GTA5$\rightarrow$Cityscape and SYNTHIA$\rightarrow$Cityscape, where GTA5~\cite{richter2016playing} and SYNTHIA~\cite{ros2016synthia} are synthetic datasets, while Cityscape~\cite{cordts2016cityscapes} is a real-world dataset. In SFSS setting, we pre-train the model on source domain dataset (GTA5 or SYNTHIA), and then adapt to the target domain dataset (Cityscapes).
		
		\subsection{Datasets}
		\textbf{GTA5} is a synthetic dataset, which contains 24966 high-resolution images collected from game video and the corresponding ground-truth segmentation map can be generated by computer graphics. We train the 19 common classes with CityScapes dataset.
		
		\textbf{SYNTHIA} is also a synthetic dataset, which contains 9400 images. And it shares 16 common classes with Cityscapes dataset.
		
		\textbf{Cityscapes} is a real-world dataset collected for  autonomous driving scenario from 50 cities around the world. It contains 2975 and 500 images for training and validation, respectively. In SFSS, the model is trained on the unlabeled training set and evaluated on the validation set with manual annotations.
				\begin{table*}[t]
					\caption{\small Results of SYNTHIA$\rightarrow$Cityscapes. The mIoU* denotes the mean IoU of 13 classes, excluding the classes with '*'. ``gain'' indicates the mIoU* improvement over using the source only under SFSS setting. We adopt the mIoU* as the evaluation metric.}
					
					\label{table:synthia}
					\vspace{-3.5mm}
					\begin{center}
						\begin{small}
							\resizebox{\textwidth}{34mm}{
								\begin{tabular}{l|c|c|cccccccccccccccc|ccc}
									\toprule[1.5pt]
									\multicolumn{22}{c}{SYNTHIA$\rightarrow$Cityscapes}\\
									\toprule
									Method &\rotatebox{90}{mech.}&SF& \rotatebox{90}{road}&\rotatebox{90}{side.}&\rotatebox{90}{build.}&\rotatebox{90}{wall*}&\rotatebox{90}{fence*}&\rotatebox{90}{pole*}&\rotatebox{90}{light}&\rotatebox{90}{sign}&\rotatebox{90}{vege.}&\rotatebox{90}{sky}&\rotatebox{90}{pers.}&\rotatebox{90}{rider}&\rotatebox{90}{car}&\rotatebox{90}{bus}&\rotatebox{90}{motor}&\rotatebox{90}{bike}&\rotatebox{90}{\bf mIoU}&\rotatebox{90}{\bf mIoU*}&\rotatebox{90}{\bf gain}\\
									\midrule
									Source only\dag&-&\multirow{9}{*}{\Checkmark}&47.1&23.3&75.5&\bf7.1&0.1&\bf23.9&5.1&9.2&74.0&73.5&51.1&20.9&39.1&17.7&18.4&34.0&32.5&37.6&-\\
									EntMin(CVPR'19)~\cite{vu2019advent}\dag& S& & 80.6&0.3&72.4&0.4&0.0&3.7&0.4&3.4&73.2&72.3&20.6&4.2&78.6&23.0&1.2&0.0&27.1&33.1&\textcolor{red}{-4.5 \%}\\
									Pseudo\dag&S& &\bf82.6&	0.0	&66.5	&0.0	&0.0	&0.3&	0.0&	0.4&	69.0&	74.5&	0.4	&0.5&	79.8&	0.4&	0.4&	0.0&23.4&28.8&\textcolor{red}{-8.8 \%}\\
									Pse.+Ent.\dag&S& &81.8	&0.0	&68.5&	0.0&	0.0&	0.3&	0.0&	0.6&	72.0&	75.1&	1.2&	0.6&	79.4&	0.4&	0.6&	0.0&	23.8&	29.2&\textcolor{red}{-8.4 \%}\\
									Pse.+Sel.\dag&S& &80.6	&2.4&	75.5&	2.2	& 0.0&	11.5&	1.1&	7.7&	75.5&	72.9&	40.0& 	9.7&	80.0&	44.1&	3.9&	1.1&	31.8&	38.0&\textcolor{green}{+0.2 \%}\\
									SHOT(ICML'20)~\cite{liang2020we}\dag& S & & 61.3	&26.4&	74.7&	5.1&	0.0&	18.8	&0.0&	\bf20.9&	75.6&	63.6&	14.5&	0.0&	52.0&	34.0&	2.2&	1.6&	28.2&	32.8&\textcolor{red}{-4.8 \%}\\
									
									LD w/o $\mathcal{L}_{neg}$(ours)\dag&S&&78.3&33.0&78.4&3.9&0.4&19.9&\bf7.3&11.4&80.0&76.8&47.5&19.0&\bf80.3&42.8&19.6&44.7&40.2&47.6&\textcolor{green}{+10.0 \%}\\
									LD w/o $\mathcal{L}_{pos}$(ours)\dag&S&&79.2&31.1&76.0&5.5&0.1&23.3&3.4&13.7&74.1&69.5&45.0&18.3&75.0&34.9&10.1&37.1&37.3&43.7&\textcolor{green}{+6.1 \%}\\
									LD(ours)\dag&S&& 77.1&	\bf33.4&	\bf79.4&	5.8&	\bf0.5&	23.7&	5.2&	13.0&	\underline{\bf81.8}&	\bf78.3&	\bf56.1&	\bf21.6&	\bf80.3&	\underline{\bf49.6}&	\bf28.0&	\bf48.1&	\bf42.6&	\bf50.1&\textcolor{green}{+12.4 \%}\\
									
									\midrule
									SIBAN(ICCV'19)~\cite{luo2019significance}& A & \multirow{8}{*}{\XSolidBrush}& 82.5& 24.0& 79.4& -& -& -& 16.5& 12.7& 79.2& 82.8& 58.3& 18.0& 79.3& 25.3& 17.6& 25.9& - & 46.3&-\\
									AdaptSeg(CVPR'18)~\cite{tsai2018learning}&A& &84.3 &42.7& 77.5&-&-&-&4.7& 7.0& 77.9& 82.5& 54.3& 21.0& 72.3& 32.2& 18.9& 32.3&-&46.7&-\\
									CLAN(PAMI'21)~\cite{luo2021category}& A& &82.7 &37.2& 81.5& - &- &-& 17.1 &13.1& 81.2& 83.3 &55.5& 22.1& 76.6& 30.1& 23.5 &30.7&-& 48.8&-\\
									DPR(ICCV'19)~\cite{tsai2019domain} & SAT& & 82.4& 38.0& 78.6& 8.7& 0.6& 26.0& 3.9& 11.1& 75.5& 84.6& 53.5& 21.6& 71.4& 32.6& 19.3& 31.7& 40.0& 46.5&-\\
									IntraDA(CVPR'20)~\cite{pan2020unsupervised}& SA& & 84.3& 37.7& 79.5& 5.3& 0.4& 24.9& 9.2& 8.4& 80.0& 84.1& 57.2& 23.0& 78.0& 38.1& 20.3& 36.5& 41.7& 48.9&-\\
									CRST(ICCV'19)~\cite{zou2019confidence}&S& & 67.7& 32.2& 73.9& 10.7& \underline{1.6}&\underline{37.4}& 22.2& \underline{31.2}& 80.8& 80.5& \underline{60.8}& \underline{29.1}& 82.8& 25.0& 19.4& 45.3& 43.8& 50.1&-\\
									DAST(AAAI'21)~\cite{yu2021dast}&SA&&\underline{87.1} &\underline{44.5}& \underline{82.3}& 10.7& 0.8& 29.9& 13.9& 13.1& 81.6& \underline{86.0}& 60.3& 25.1& \underline{83.1}& 40.1& 24.4& 40.5& \underline{45.2}& 52.5&-\\
									
									CCM(ECCV'20)~\cite{li2020content} &S&&79.6 &36.4&80.6&\underline{13.3}&0.3&25.5& \underline{22.4}&14.9&\underline{81.8}&77.4&56.8&25.9&80.7&45.3&\underline{29.9}&\underline{52.0}&\underline{45.2}&\underline{52.9}&-\\
									\bottomrule[1.5pt]
								\end{tabular}}
							\end{small}
						\end{center}
					\end{table*}
		\subsection{Implementation Details}
		We keep the same network architecture as in previous methods for fair comparison. Specifically, we use Deeplab-V2~\cite{chen2017deeplab} with pre-trained ResNet-101~\cite{he2016deep} as the segmentation network $S$ and the Atrous Spatial Pyramid Pooling (ASPP) module is applied on the last layer's output. Similar to ~\cite{chen2017deeplab}, the sampling rates are fixed as \{6,12,18,24\}. Then, an up-sampling layer and a softmax operator are applied to obtain the prediction (segmentation map) with the matched size of input image. We implement our methods with Pytorch on a single NVIDIA RTX 2080Ti GPU. The segmentation network $S$ is trained using the Stochastic Gradient Descent optimizer with momentum 0.9 and weight decay $5\times10^{-4}$. We adopt the polynomial learning rate scheduling with the power of 0.9, the initial learning rate is set as $1\times10^{-3}$ and the batch size is 3. We adopt the same data augmentations as~\cite{li2020content}: we resize the images' short side to 720 and crop the images into 600$\times$600 randomly. And horizontal flip and random scale between 0.5 and 1.5 are performed. The evaluation metrics is the mean intersection-over-union (mIoU).
		
		\subsection{Evaluation Protocols and Baselines}
		\label{sec:methods}
		To evaluate the proposed LD, we compare it with some widely-used self-training methods in domain adaptive semantic segmentation~\cite{vu2019advent} and the state-of-the-arts SFDA method SHOT~\cite{liang2020we}. Here, we shortly introduce the compared methods and our implementation since some methods cannot be directly applied to source-free domain adaptive semantic segmentation.
		\textbf{EntMin.}  We minimize the entropy of softmax output~\cite{vu2019advent}.
		\textbf{Pseudo.} We directly obtain the pseudo label with argmax operation. Then, we minimize the cross-entropy loss on each pixel~\cite{lee2013pseudo} .
		\textbf{Pse.+Ent.} We combine the aforementioned two methods, and the trade-off hyper-parameter is set as 1.
		\textbf{Pse.+Sel.} Since directly applying the pseudo label leads to many noisy predictions, many methods proposed various selection strategies~\cite{zheng2021rectifying,zou2019confidence,zou2018unsupervised}. A recent work~\cite{li2020free}, which investigates the SFDA in object detection, proposed to generate the pseudo labels through prediction confidence. In our implementation, we empirically fix the confidence threshold as 0.9, and minimize the cross-entropy loss on the selected pixels.
		\textbf{SHOT.} SHOT contains information maximization and deep clustering. However, the latter one cannot be applied to the semantic segmentation scenario directly. For instance, in semantic segmentation, a 600$\times$600 image contains 360000 pixels with 2048-dim (ResNet) feature representations and a dataset contains large amounts of images. Therefore, performing clustering on the dataset in semantic segmentation is impractical and we implement information maximization.
			\begin{figure*}[t]
				\centering
				\includegraphics[width=\linewidth]{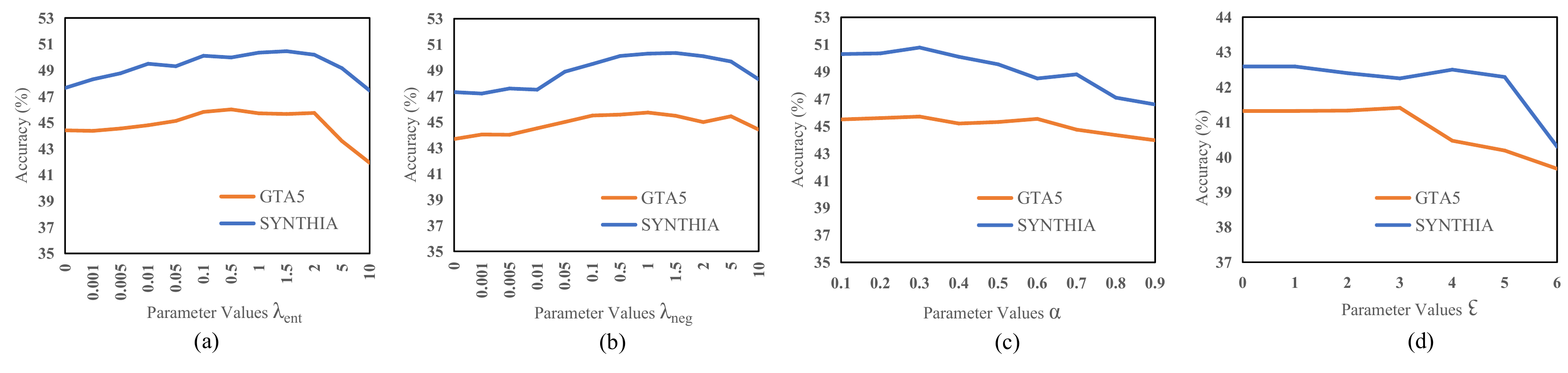}
				\vskip -15pt
				\caption{\small (Best viewed in color.) The results of parameter sensitivity on GTA5$\rightarrow$Cityscapes and SYNTHIA$\rightarrow$Cityscapes. (a) The weight of entropy minimization loss. (b) The weight of negative learning loss. (c) The selection range $\alpha$ in positive learning. (d) The selection range $\epsilon$ in negative learning. }
				\label{fig:param}
				\vskip -8pt
			\end{figure*}
			\subsection{Results}
			\label{sec:result}
			We verify the effectiveness of our method in two synthetic-to-real benchmarks: GTA5 $\rightarrow$ Cityscapes and SYNTHIA$\rightarrow$Cityscapes. To ensure a fair comparison, all reported results share the same backbone: DeepLab-v2 with pre-trained ResNet-101. Since existing DASS methods cannot be implemented directly in the source-free settings, we then re-implement some widely-used self-training approaches in DASS~\cite{vu2019advent} and the state-of-the-art SFDA method SHOT~\cite{liang2020we}. The details of implementation are elaborated in Sec.~\ref{sec:methods}.
			The results on GTA5$\rightarrow$Cityscapes and SYNTHIA$\rightarrow$Cityscapes are shown in Table \ref{table:gta5} and Table \ref{table:synthia}, respectively. From Table \ref{table:gta5} and \ref{table:synthia}, we have following findings:
			\begin{enumerate}[1)]
				\item Entropy minimization and direct pseudo-labeling leads to ``winner-takes-all'' under SFSS setting. Compare to Source-only, it has $9.7\%$, $12.3\%$ drop in GTA5$\rightarrow$Cityscapes, and $4.5\%$, $8.8\%$ drop in SYNTHIA$\rightarrow$Cityscapes. Both in DASS and SFSS, the model is biased towards the {\it majority} classes. However, in DASS, the labeled source domain data preserve the {\it minority} classes, while in SFSS, once the class is at a disadvantage, it will be gradually replaced if there is no external intervention. We illustrate the results of SYNTHIA$\rightarrow$Cityscapes in Fig.~\ref{fig:fig2}, which vividly demonstrates our claims.
				\item Selecting the pseudo-labeled samples based on the confidence threshold (Pse.+Sel.) effectively removes the wrongly pseudo-labeled samples, but still suffers from the ``winner-takes-all''. Pse.+Sel. gets $-3.2\%$ in GTA5$\rightarrow$Cityscapes, and $+0.3\%$ in SYNTHIA. However, in the {\it minority} classes (e.g., fence, pole, light, sign and etc.), Pse.+Sel. has a clear drop compared to ``Source only'', revealing that the selected pixels are extremely class-imbalanced and the segmentation networks suffer from the ``winner-tasks-all'' dilemma.
				Existing state-of-the-art SFDA approach SHOT also gets poor performance in semantic segmentation scenario (i.e., +2.8\% mIoU in GTA5$\rightarrow$Cityscapes and -4.8\% mIoU in SYNTHIA$\rightarrow$Cityscapes). 
				\item Our approach LD surpasses the strong baseline ``Source only'' by a large margin under the challenging SFSS setting. We achieve a promising mIoU of 45.5\% in GTA5$\rightarrow$Cityscapes (+9.8\%) and mIoU of 50.1\% in SYNTHIA$\rightarrow$Cityscapes (+12.4\%), which highlights the effectiveness of our proposed LD. Compared to other methods under SFSS setting, our method maintains (e.g., wall, fence, pole, light, sign and etc.) or upgrades (e.g., side., rider, truck, bus and etc.) the performance of {\it minority} classes and consistently improves the performance of {\it majority} classes. 
				\item Our LD even achieves competitive performance compare to the DASS methods, which access the labeled source domain data during adapting to the target domain. Our method outperforms the recent state-of-the-art adversarial based methods. For example, we surpass the representative adversarial-based work CLAN with a 1.3\% mIoU in SYNTHIA$\rightarrow$Cityscapes. Compared to state-of-the-art self-training based methods, our performance is also competitive. Notably, in SYNTHIA$\rightarrow$Cityscapes, we achieves the best or the second performance at classes ``bus'', ``vege.'', ``motor'' and ``bike'', and all of them belong to the {\it minority} classes except for ``vege.''. Furthermore, our method only applies two self-training components, while many DASS methods combine many methods and design complex network architectures or sampling strategies (e.g., DPR, IntraDA and DAST). i.e., incorporating with other methods (e.g., attention module) will further enhance the performance of LD, which revealing the simplicity and expandability of the proposed framework. 
			\end{enumerate}

			\section{Analysis}
			
			\subsection{Parameter Sensitivity}
			\textbf{Hyper-parameters $\lambda_{ent}$, $\lambda_{neg}$.}
			The overall objective function Eq.~(\ref{eq:overall}) contains two trade-off hyper-parameters $\lambda_{ent}$ and $\lambda_{neg}$. We fix $\lambda_{ent}$ as 1 and $\lambda_{neg}$ as 1 for all tasks, respectively. Furthermore, we conduct parameter sensitivity analysis to evaluate LD on tasks GTA5$\rightarrow$Cityscapes and SYNTHIA$\rightarrow$Cityscapes. As shown in Fig.~\ref{fig:param} (a) and (b), the performance steadily improves with the increasing parameters $\lambda_{ent}$ and $\lambda_{neg}$ from 0 to 1, demonstrating the effectiveness of each components in LD. And we observe that the performance would not be greatly influenced by the value of $\lambda_{ent}$ and $\lambda_{neg}$, indicating that our LD is not quite sensitive to $\lambda_{ent}$ and $\lambda_{neg}$.
			
			\textbf{Positive learning: selection range $\alpha$.}
			\label{sec:posrange}
			In Eq.~(\ref{eq:alpha}), the parameter $\alpha$ controls the selection range of pseudo labels and we fix $\alpha$ as 0.2 for all tasks. To fully investigate the influence of different value of $\alpha$, we select the balance weights from 0.1 to 0.9, since $\alpha=0$ indicates applying negative learning only and $\alpha=1$ indicates applying pseudo-labeling directly. From Fig.~\ref{fig:param} (c), we find that the performance is not affected much by different values of $\alpha$ in GTA5$\rightarrow$Cityscapes. In the more challenging task SYNTHIA$\rightarrow$Cityscapes, the performance begins to drop after $\alpha >0.3$ since it contains more and more wrong labels. It is worth noting that even with a large value $\alpha$, the performance of LD is also acceptable, indicating the negative learning is effective in learning with noisy labels.
			
			\begin{figure*}[t]
				\centering
				\includegraphics[width=\linewidth]{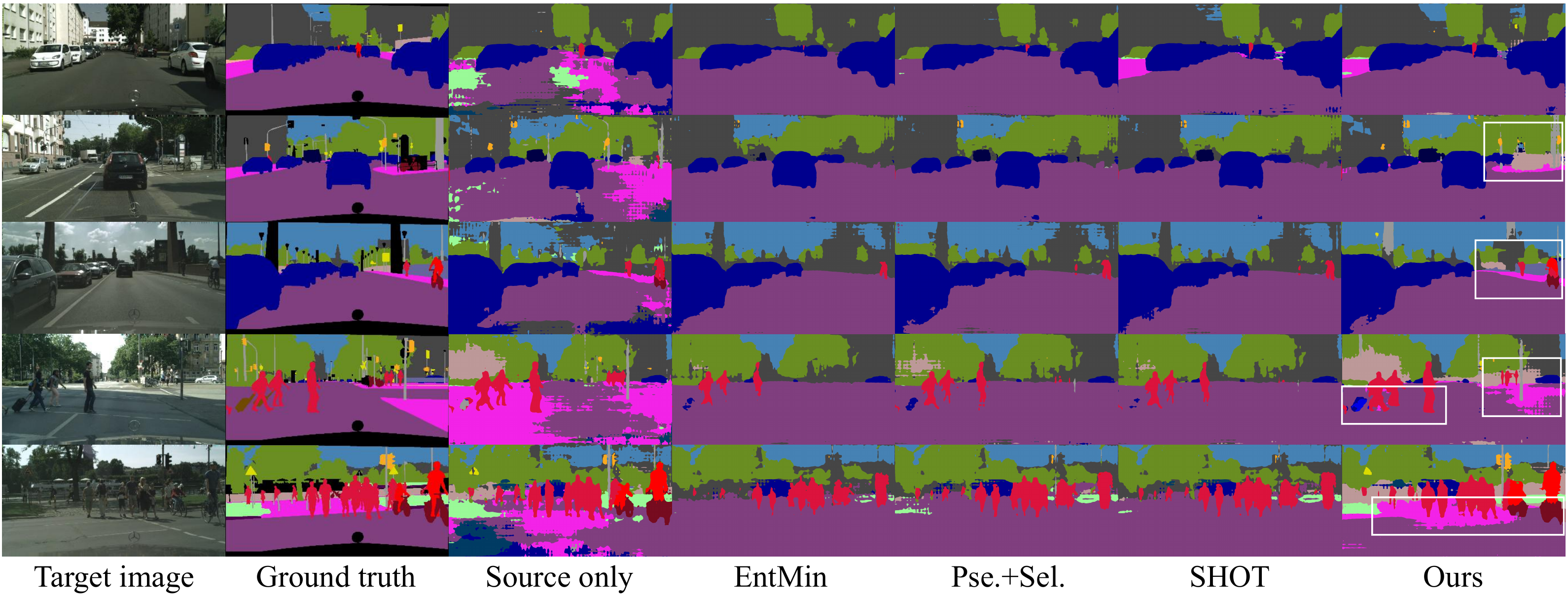}
				\vskip -10pt
				\caption{\small (Best viewed in color.) Qualitative results of source-free domain adaptive semantic segmentation on GTA5$\rightarrow$Cityscapes. The key areas are highlighted with the white boxes.}
				\label{fig:qua}
			\end{figure*}
			
			\textbf{Negative learning: selection range $\epsilon$.}
			In Algorithm~\ref{alg:hcls}, $\epsilon$ represents the selection width of complementary label and we fix $\epsilon=3$ for all tasks. To better reflect the influence of different value of $\epsilon$, we apply negative learning only and modify the Eq.~(\ref{eq:epsilon}) to $K= \lfloor C/2 \rfloor$ + rand ($-\epsilon,0$). In GTA5$\rightarrow$Cityscapes, the number of classes $C=19$, and $\lfloor C/2 \rfloor=9$, while in SYNTHIA$\rightarrow$Cityscapes, $C=16$ and $K= \lfloor C/2 \rfloor=8$. Therefore, we select the balance weights from 0 to 6, since $\epsilon=7$ indicates that the complementary label may have the Top 1 value of the softmax output in SYNTHIA$\rightarrow$Cityscapes. As shown in Fig.~\ref{fig:param} (d), we observe that the performance begins to drop when $\epsilon >=4$ in GTA5$\rightarrow$Cityscapes, which is consistent with our claims in Sec.~\ref{sec:neg} and Fig.~\ref{fig:fig2}: generating the complementary labels from the ``head'' classes results in the degeneration of model. Specifically, the empirical results indicate that our method outperforms the mainstream complementary strategy~\cite{ishida2017learning,kim2019nlnl}, who generating the complementary label randomly from the labels of all classes except for the pseudo label, since our method generates more accurate and reliable complementary labels to perform negative learning. 
			\vspace{-2mm}
			\subsection{Ablation Study}
			For the sake of examining the key components of LD, we perform ablation study by removing each component from LD at a time. In LD, we have two simple components: positive learning and negative learning. Therefore, we obtain two variants: (1) ``LD w/o $\mathcal{L}_{pos}$'' and (2) ``LD w/o $\mathcal{L}_{neg}$'', which denote removing the corresponding self-training module, respectively. The results of ablation study are shown in Table~\ref{table:gta5} and~\ref{table:synthia}. It is obvious that LD outperforms other variants and achieves significant improvement under source-free setting. ``LD w/o $\mathcal{L}_{pos}$'' achieves the mIoU of 41.3 in GTA5$\rightarrow$Cityscapes and 43.7 in SYNTHIA$\rightarrow$Cityscapes. ``LD w/o $\mathcal{L}_{pos}$'' does not explicitly provide any supervision but tells the model ``which category does this pixel not belong to''. ``LD w/o $\mathcal{L}_{neg}$'' achieves the mIoU of 43.0 in GTA5$\rightarrow$Cityscapes and 47.6 in SYNTHIA$\rightarrow$Cityscapes, revealing that our pseudo-labeling effectively alleviate the ``winner-takes-all'' and enhance the performance of model in source-free setting.
			\vspace{-2mm}
			\subsection{Qualitative Analysis}
			We illustrate some qualitative segmentation examples in Fig.~\ref{fig:qua}. Obviously, ``Source only'' model gives confused predictions due to the domain gap between GTA5 and Cityscapes. And many self-training methods performs well on the {\it majority} classes (e.g., ``road'', ``car''), but tends to over-fit the {\it majority} classes, e.g., ``road'' tends to occupy the ``sidewalk''. Furthermore, they totally ignore the {\it minority} classes. For instance, the ``bike'' class is totally disappeared and replaced with ``road'' class, which creates a potential risk for real-world applications like automatic pilot. Unlike them, the proposed LD significantly enhances the performance of semantic segmentation. Specifically, LD stably improves the performance of {\it majority} classes, making the segmentation cleaner compared to ``Source only'' model. Furthermore, from the key areas highlighted with white boxes, our method yields better scalability to the {\it minority} classes, like confused classes (e.g., ``bike'') and small-scale objectives (e.g., ``traffic sign''). The qualitative results vividly confirms our claims and demonstrates the superiority of our method.
	
				\subsection{Efficiency Validation}
				\label{sec:eff}
				Without accessing source data, our method shows superiority on privacy-preserving, data transmission and training time. In this section, we directly measure the size of training data and the training time of our proposed LD and other methods on task GTA5$\rightarrow$Cityscapes and SYNTHIA$\rightarrow$Cityscapes. We run the experiments on the same running environment (same hardware, same OS, same background applications, etc.). The results are reported in Table~\ref{table:eff}. In Table~\ref{table:eff}, ``Training data'' indicates the size of data when adapting to the target domain. ``Training time'' represents the time taken by the model to reach the reported mIoU. ``Speedup'' is calculated by: (The training time of the compared method - The training time of LD)/(The training time of LD). From the results, we can find that our method only takes 6.9GB training data during adapting to the target domain and has a significant speedup compared to the source-accessible methods. Meanwhile, our performance is also competitive to these methods, and even better than CLAN and IntraDA on SYNTHIA$\rightarrow$Cityscapes. Therefore, our method achieves efficient domain adaptation.
						\begin{table}[t]
							\vskip -7pt
							\caption{\small The results of efficiency validation.}
							
							\label{table:eff}
							
							\begin{center}
								\begin{small}
									{\vskip -7pt
										\begin{tabular}{lccc}
											\toprule
											Method &Training data& Training time &Speedup\\
											\midrule
											\multicolumn{4}{c}{GTA5$\rightarrow$Cityscapes}\\
											\midrule
											
											CLAN(PAMI'21)~\cite{luo2021category} &68.5GB&36572.32s&12.33$\times$\\
											IntraDA(CVPR'20)~\cite{pan2020unsupervised}&68.5GB&43531.11s&14.67$\times$\\
											CCM(ECCV'20)~\cite{li2020content}&68.5GB&29285.02s&9.86$\times$\\
											LD(ours)&6.9GB&2967.85s&-\\
											\midrule
											\multicolumn{4}{c}{SYNTHIA$\rightarrow$Cityscapes}\\
											\midrule
											CLAN(PAMI'21)~\cite{luo2021category} &27.8GB & 16390.57s& 5.00$\times$\\
											IntraDA(CVPR'20)~\cite{pan2020unsupervised}&27.8GB&31017.96s&9.45$\times$\\
											CCM(ECCV'20)~\cite{li2020content}&27.8GB&18196.11s&5.72$\times$\\
											LD(ours)&6.9GB&3281.14s&-\\
											\bottomrule
										\end{tabular}}
										\vspace{-2mm}
									\end{small}
								\end{center}
							\end{table}
							\vspace{-4.5mm}
				\section{Conclusion}
				In this paper, we propose an effective and efficient framework named LD for source-free domain adaptive semantic segmentation. Specifically, LD selects the class-balanced pseudo-labeled pixels and performs negative learning with the proposed HCLS. Extensive experiments verify that LD outperforms the baseline with a large margin and even achieves competitive performance compared to existing state-of-the-art methods, which access the source data when adapting to the target domain. It is worth noting that our method can be easily incorporated with other modules to achieve better performance. We hope our framework can provide a solid baseline in source-free domain adaptive semantic segmentation and bring inspiration for future research. 
				
				\section*{Acknowledgement}
				This work was supported in part by the National Natural Science Foundation of China under Grant 61806039 and 62073059, and in part by Sichuan Science and Technology Program under Grant 2020YFG0080 and 2020YFG0481. 
				
				\bibliographystyle{ACM-Reference-Format}
				\bibliography{sample-base}


\begin{thebibliography}{53}


\ifx \showCODEN    \undefined \def \showCODEN     #1{\unskip}     \fi
\ifx \showDOI      \undefined \def \showDOI       #1{#1}\fi
\ifx \showISBNx    \undefined \def \showISBNx     #1{\unskip}     \fi
\ifx \showISBNxiii \undefined \def \showISBNxiii  #1{\unskip}     \fi
\ifx \showISSN     \undefined \def \showISSN      #1{\unskip}     \fi
\ifx \showLCCN     \undefined \def \showLCCN      #1{\unskip}     \fi
\ifx \shownote     \undefined \def \shownote      #1{#1}          \fi
\ifx \showarticletitle \undefined \def \showarticletitle #1{#1}   \fi
\ifx \showURL      \undefined \def \showURL       {\relax}        \fi
\providecommand\bibfield[2]{#2}
\providecommand\bibinfo[2]{#2}
\providecommand\natexlab[1]{#1}
\providecommand\showeprint[2][]{arXiv:#2}

\bibitem[\protect\citeauthoryear{Ben-David, Blitzer, Crammer, Kulesza, Pereira,
  and Vaughan}{Ben-David et~al\mbox{.}}{2010}]%
        {ben2010theory}
\bibfield{author}{\bibinfo{person}{Shai Ben-David}, \bibinfo{person}{John
  Blitzer}, \bibinfo{person}{Koby Crammer}, \bibinfo{person}{Alex Kulesza},
  \bibinfo{person}{Fernando Pereira}, {and} \bibinfo{person}{Jennifer~Wortman
  Vaughan}.} \bibinfo{year}{2010}\natexlab{}.
\newblock \showarticletitle{A theory of learning from different domains}.
\newblock \bibinfo{journal}{\emph{Machine learning}} \bibinfo{volume}{79},
  \bibinfo{number}{1-2} (\bibinfo{year}{2010}), \bibinfo{pages}{151--175}.
\newblock


\bibitem[\protect\citeauthoryear{Chang, Wang, Peng, and Chiu}{Chang
  et~al\mbox{.}}{2019}]%
        {chang2019all}
\bibfield{author}{\bibinfo{person}{Wei-Lun Chang}, \bibinfo{person}{Hui-Po
  Wang}, \bibinfo{person}{Wen-Hsiao Peng}, {and} \bibinfo{person}{Wei-Chen
  Chiu}.} \bibinfo{year}{2019}\natexlab{}.
\newblock \showarticletitle{All about structure: Adapting structural
  information across domains for boosting semantic segmentation}. In
  \bibinfo{booktitle}{\emph{Proceedings of the IEEE/CVF Conference on Computer
  Vision and Pattern Recognition}}. \bibinfo{pages}{1900--1909}.
\newblock


\bibitem[\protect\citeauthoryear{Chen, Papandreou, Kokkinos, Murphy, and
  Yuille}{Chen et~al\mbox{.}}{2017}]%
        {chen2017deeplab}
\bibfield{author}{\bibinfo{person}{Liang-Chieh Chen}, \bibinfo{person}{George
  Papandreou}, \bibinfo{person}{Iasonas Kokkinos}, \bibinfo{person}{Kevin
  Murphy}, {and} \bibinfo{person}{Alan~L Yuille}.}
  \bibinfo{year}{2017}\natexlab{}.
\newblock \showarticletitle{Deeplab: Semantic image segmentation with deep
  convolutional nets, atrous convolution, and fully connected crfs}.
\newblock \bibinfo{journal}{\emph{IEEE transactions on pattern analysis and
  machine intelligence}} \bibinfo{volume}{40}, \bibinfo{number}{4}
  (\bibinfo{year}{2017}), \bibinfo{pages}{834--848}.
\newblock


\bibitem[\protect\citeauthoryear{Chen, Xue, and Cai}{Chen
  et~al\mbox{.}}{2019b}]%
        {chen2019domain}
\bibfield{author}{\bibinfo{person}{Minghao Chen}, \bibinfo{person}{Hongyang
  Xue}, {and} \bibinfo{person}{Deng Cai}.} \bibinfo{year}{2019}\natexlab{b}.
\newblock \showarticletitle{Domain adaptation for semantic segmentation with
  maximum squares loss}. In \bibinfo{booktitle}{\emph{Proceedings of the
  IEEE/CVF International Conference on Computer Vision}}.
  \bibinfo{pages}{2090--2099}.
\newblock


\bibitem[\protect\citeauthoryear{Chen, Lin, Yang, and Huang}{Chen
  et~al\mbox{.}}{2019a}]%
        {chen2019crdoco}
\bibfield{author}{\bibinfo{person}{Yun-Chun Chen}, \bibinfo{person}{Yen-Yu
  Lin}, \bibinfo{person}{Ming-Hsuan Yang}, {and} \bibinfo{person}{Jia-Bin
  Huang}.} \bibinfo{year}{2019}\natexlab{a}.
\newblock \showarticletitle{Crdoco: Pixel-level domain transfer with
  cross-domain consistency}. In \bibinfo{booktitle}{\emph{Proceedings of the
  IEEE/CVF Conference on Computer Vision and Pattern Recognition}}.
  \bibinfo{pages}{1791--1800}.
\newblock


\bibitem[\protect\citeauthoryear{Cordts, Omran, Ramos, Rehfeld, Enzweiler,
  Benenson, Franke, Roth, and Schiele}{Cordts et~al\mbox{.}}{2016}]%
        {cordts2016cityscapes}
\bibfield{author}{\bibinfo{person}{Marius Cordts}, \bibinfo{person}{Mohamed
  Omran}, \bibinfo{person}{Sebastian Ramos}, \bibinfo{person}{Timo Rehfeld},
  \bibinfo{person}{Markus Enzweiler}, \bibinfo{person}{Rodrigo Benenson},
  \bibinfo{person}{Uwe Franke}, \bibinfo{person}{Stefan Roth}, {and}
  \bibinfo{person}{Bernt Schiele}.} \bibinfo{year}{2016}\natexlab{}.
\newblock \showarticletitle{The cityscapes dataset for semantic urban scene
  understanding}. In \bibinfo{booktitle}{\emph{Proceedings of the IEEE
  conference on computer vision and pattern recognition}}.
  \bibinfo{pages}{3213--3223}.
\newblock


\bibitem[\protect\citeauthoryear{Grandvalet, Bengio, et~al\mbox{.}}{Grandvalet
  et~al\mbox{.}}{2005}]%
        {grandvalet2005semi}
\bibfield{author}{\bibinfo{person}{Yves Grandvalet}, \bibinfo{person}{Yoshua
  Bengio}, {et~al\mbox{.}}} \bibinfo{year}{2005}\natexlab{}.
\newblock \showarticletitle{Semi-supervised learning by entropy minimization.}.
  In \bibinfo{booktitle}{\emph{CAP}}. \bibinfo{pages}{281--296}.
\newblock


\bibitem[\protect\citeauthoryear{Guo, Liu, Tang, Li, Luo, and Lu}{Guo
  et~al\mbox{.}}{2019}]%
        {guo2019aligning}
\bibfield{author}{\bibinfo{person}{Longteng Guo}, \bibinfo{person}{Jing Liu},
  \bibinfo{person}{Jinhui Tang}, \bibinfo{person}{Jiangwei Li},
  \bibinfo{person}{Wei Luo}, {and} \bibinfo{person}{Hanqing Lu}.}
  \bibinfo{year}{2019}\natexlab{}.
\newblock \showarticletitle{Aligning linguistic words and visual semantic units
  for image captioning}. In \bibinfo{booktitle}{\emph{Proceedings of the 27th
  ACM International Conference on Multimedia}}. \bibinfo{pages}{765--773}.
\newblock


\bibitem[\protect\citeauthoryear{He, Zhang, Ren, and Sun}{He
  et~al\mbox{.}}{2016}]%
        {he2016deep}
\bibfield{author}{\bibinfo{person}{Kaiming He}, \bibinfo{person}{Xiangyu
  Zhang}, \bibinfo{person}{Shaoqing Ren}, {and} \bibinfo{person}{Jian Sun}.}
  \bibinfo{year}{2016}\natexlab{}.
\newblock \showarticletitle{Deep residual learning for image recognition}. In
  \bibinfo{booktitle}{\emph{Proceedings of the IEEE conference on computer
  vision and pattern recognition}}. \bibinfo{pages}{770--778}.
\newblock


\bibitem[\protect\citeauthoryear{Hoffman, Tzeng, Park, Zhu, Isola, Saenko,
  Efros, and Darrell}{Hoffman et~al\mbox{.}}{2017}]%
        {hoffman2017cycada}
\bibfield{author}{\bibinfo{person}{Judy Hoffman}, \bibinfo{person}{Eric Tzeng},
  \bibinfo{person}{Taesung Park}, \bibinfo{person}{Jun-Yan Zhu},
  \bibinfo{person}{Phillip Isola}, \bibinfo{person}{Kate Saenko},
  \bibinfo{person}{Alexei~A Efros}, {and} \bibinfo{person}{Trevor Darrell}.}
  \bibinfo{year}{2017}\natexlab{}.
\newblock \showarticletitle{Cycada: Cycle-consistent adversarial domain
  adaptation}.
\newblock \bibinfo{journal}{\emph{arXiv preprint arXiv:1711.03213}}
  (\bibinfo{year}{2017}).
\newblock


\bibitem[\protect\citeauthoryear{Hou and Zheng}{Hou and Zheng}{2020}]%
        {hou2020source}
\bibfield{author}{\bibinfo{person}{Yunzhong Hou} {and} \bibinfo{person}{Liang
  Zheng}.} \bibinfo{year}{2020}\natexlab{}.
\newblock \showarticletitle{Source Free Domain Adaptation with Image
  Translation}.
\newblock \bibinfo{journal}{\emph{arXiv preprint arXiv:2008.07514}}
  (\bibinfo{year}{2020}).
\newblock


\bibitem[\protect\citeauthoryear{Iqbal and Ali}{Iqbal and Ali}{2020}]%
        {iqbal2020mlsl}
\bibfield{author}{\bibinfo{person}{Javed Iqbal} {and} \bibinfo{person}{Mohsen
  Ali}.} \bibinfo{year}{2020}\natexlab{}.
\newblock \showarticletitle{Mlsl: Multi-level self-supervised learning for
  domain adaptation with spatially independent and semantically consistent
  labeling}. In \bibinfo{booktitle}{\emph{Proceedings of the IEEE/CVF Winter
  Conference on Applications of Computer Vision}}. \bibinfo{pages}{1864--1873}.
\newblock


\bibitem[\protect\citeauthoryear{Ishida, Niu, Hu, and Sugiyama}{Ishida
  et~al\mbox{.}}{2017}]%
        {ishida2017learning}
\bibfield{author}{\bibinfo{person}{Takashi Ishida}, \bibinfo{person}{Gang Niu},
  \bibinfo{person}{Weihua Hu}, {and} \bibinfo{person}{Masashi Sugiyama}.}
  \bibinfo{year}{2017}\natexlab{}.
\newblock \showarticletitle{Learning from complementary labels}.
\newblock \bibinfo{journal}{\emph{arXiv preprint arXiv:1705.07541}}
  (\bibinfo{year}{2017}).
\newblock


\bibitem[\protect\citeauthoryear{Kim, Hong, Cho, Park, and Panda}{Kim
  et~al\mbox{.}}{2020}]%
        {kim2020domain}
\bibfield{author}{\bibinfo{person}{Youngeun Kim}, \bibinfo{person}{Sungeun
  Hong}, \bibinfo{person}{Donghyeon Cho}, \bibinfo{person}{Hyoungseob Park},
  {and} \bibinfo{person}{Priyadarshini Panda}.}
  \bibinfo{year}{2020}\natexlab{}.
\newblock \showarticletitle{Domain Adaptation without Source Data}.
\newblock \bibinfo{journal}{\emph{arXiv preprint arXiv:2007.01524}}
  (\bibinfo{year}{2020}).
\newblock


\bibitem[\protect\citeauthoryear{Kim, Yim, Yun, and Kim}{Kim
  et~al\mbox{.}}{2019}]%
        {kim2019nlnl}
\bibfield{author}{\bibinfo{person}{Youngdong Kim}, \bibinfo{person}{Junho Yim},
  \bibinfo{person}{Juseung Yun}, {and} \bibinfo{person}{Junmo Kim}.}
  \bibinfo{year}{2019}\natexlab{}.
\newblock \showarticletitle{Nlnl: Negative learning for noisy labels}. In
  \bibinfo{booktitle}{\emph{Proceedings of the IEEE/CVF International
  Conference on Computer Vision}}. \bibinfo{pages}{101--110}.
\newblock


\bibitem[\protect\citeauthoryear{Lee et~al\mbox{.}}{Lee et~al\mbox{.}}{2013}]%
        {lee2013pseudo}
\bibfield{author}{\bibinfo{person}{Dong-Hyun Lee} {et~al\mbox{.}}}
  \bibinfo{year}{2013}\natexlab{}.
\newblock \showarticletitle{Pseudo-label: The simple and efficient
  semi-supervised learning method for deep neural networks}. In
  \bibinfo{booktitle}{\emph{Workshop on challenges in representation learning,
  ICML}}, Vol.~\bibinfo{volume}{3}.
\newblock


\bibitem[\protect\citeauthoryear{Li, Kang, Liu, Wei, and Yang}{Li
  et~al\mbox{.}}{2020d}]%
        {li2020content}
\bibfield{author}{\bibinfo{person}{Guangrui Li}, \bibinfo{person}{Guoliang
  Kang}, \bibinfo{person}{Wu Liu}, \bibinfo{person}{Yunchao Wei}, {and}
  \bibinfo{person}{Yi Yang}.} \bibinfo{year}{2020}\natexlab{d}.
\newblock \showarticletitle{Content-consistent matching for domain adaptive
  semantic segmentation}. In \bibinfo{booktitle}{\emph{European Conference on
  Computer Vision}}. Springer, \bibinfo{pages}{440--456}.
\newblock


\bibitem[\protect\citeauthoryear{Li, Chen, Ding, Zhu, Lu, and Huang}{Li
  et~al\mbox{.}}{2019a}]%
        {li2019cycle}
\bibfield{author}{\bibinfo{person}{Jingjing Li}, \bibinfo{person}{Erpeng Chen},
  \bibinfo{person}{Zhengming Ding}, \bibinfo{person}{Lei Zhu},
  \bibinfo{person}{Ke Lu}, {and} \bibinfo{person}{Zi Huang}.}
  \bibinfo{year}{2019}\natexlab{a}.
\newblock \showarticletitle{Cycle-consistent conditional adversarial transfer
  networks}. In \bibinfo{booktitle}{\emph{Proceedings of the 27th ACM
  International Conference on Multimedia}}. \bibinfo{pages}{747--755}.
\newblock


\bibitem[\protect\citeauthoryear{Li, Chen, Ding, Zhu, Lu, and Shen}{Li
  et~al\mbox{.}}{2020a}]%
        {li2020maximum}
\bibfield{author}{\bibinfo{person}{Jingjing Li}, \bibinfo{person}{Erpeng Chen},
  \bibinfo{person}{Zhengming Ding}, \bibinfo{person}{Lei Zhu},
  \bibinfo{person}{Ke Lu}, {and} \bibinfo{person}{Heng~Tao Shen}.}
  \bibinfo{year}{2020}\natexlab{a}.
\newblock \showarticletitle{Maximum Density Divergence for Domain Adaptation}.
\newblock \bibinfo{journal}{\emph{IEEE Transactions on Pattern Analysis and
  Machine Intelligence}} (\bibinfo{year}{2020}).
\newblock


\bibitem[\protect\citeauthoryear{Li, Jing, Lu, Zhu, and Shen}{Li
  et~al\mbox{.}}{2019b}]%
        {li2019locality}
\bibfield{author}{\bibinfo{person}{Jingjing Li}, \bibinfo{person}{Mengmeng
  Jing}, \bibinfo{person}{Ke Lu}, \bibinfo{person}{Lei Zhu}, {and}
  \bibinfo{person}{Heng~Tao Shen}.} \bibinfo{year}{2019}\natexlab{b}.
\newblock \showarticletitle{Locality preserving joint transfer for domain
  adaptation}.
\newblock \bibinfo{journal}{\emph{IEEE Transactions on Image Processing}}
  \bibinfo{volume}{28}, \bibinfo{number}{12} (\bibinfo{year}{2019}),
  \bibinfo{pages}{6103--6115}.
\newblock


\bibitem[\protect\citeauthoryear{Li, Jing, Su, Lu, Zhu, and Shen}{Li
  et~al\mbox{.}}{2021}]%
        {li2021faster}
\bibfield{author}{\bibinfo{person}{Jingjing Li}, \bibinfo{person}{Mengmeng
  Jing}, \bibinfo{person}{Hongzu Su}, \bibinfo{person}{Ke Lu},
  \bibinfo{person}{Lei Zhu}, {and} \bibinfo{person}{Heng~Tao Shen}.}
  \bibinfo{year}{2021}\natexlab{}.
\newblock \showarticletitle{Faster domain adaptation networks}.
\newblock \bibinfo{journal}{\emph{IEEE Transactions on Knowledge and Data
  Engineering}} (\bibinfo{year}{2021}).
\newblock


\bibitem[\protect\citeauthoryear{Li, Lu, Huang, Zhu, and Shen}{Li
  et~al\mbox{.}}{2018}]%
        {li2018heterogeneous}
\bibfield{author}{\bibinfo{person}{Jingjing Li}, \bibinfo{person}{Ke Lu},
  \bibinfo{person}{Zi Huang}, \bibinfo{person}{Lei Zhu}, {and}
  \bibinfo{person}{Heng~Tao Shen}.} \bibinfo{year}{2018}\natexlab{}.
\newblock \showarticletitle{Heterogeneous domain adaptation through progressive
  alignment}.
\newblock \bibinfo{journal}{\emph{IEEE transactions on neural networks and
  learning systems}} \bibinfo{volume}{30}, \bibinfo{number}{5}
  (\bibinfo{year}{2018}), \bibinfo{pages}{1381--1391}.
\newblock


\bibitem[\protect\citeauthoryear{Li, Lu, Huang, Zhu, and Shen}{Li
  et~al\mbox{.}}{2019c}]%
        {li2019transfer}
\bibfield{author}{\bibinfo{person}{Jingjing Li}, \bibinfo{person}{Ke Lu},
  \bibinfo{person}{Zi Huang}, \bibinfo{person}{Lei Zhu}, {and}
  \bibinfo{person}{Heng~Tao Shen}.} \bibinfo{year}{2019}\natexlab{c}.
\newblock \showarticletitle{Transfer Independently Together: A Generalized
  Framework for Domain Adaptation}.
\newblock \bibinfo{journal}{\emph{IEEE Transactions on Cybernetics}}
  (\bibinfo{year}{2019}).
\newblock


\bibitem[\protect\citeauthoryear{Li, Jiao, Cao, Wong, and Wu}{Li
  et~al\mbox{.}}{2020c}]%
        {li2020model}
\bibfield{author}{\bibinfo{person}{Rui Li}, \bibinfo{person}{Qianfen Jiao},
  \bibinfo{person}{Wenming Cao}, \bibinfo{person}{Hau-San Wong}, {and}
  \bibinfo{person}{Si Wu}.} \bibinfo{year}{2020}\natexlab{c}.
\newblock \showarticletitle{Model adaptation: Unsupervised domain adaptation
  without source data}. In \bibinfo{booktitle}{\emph{Proceedings of the
  IEEE/CVF Conference on Computer Vision and Pattern Recognition}}.
  \bibinfo{pages}{9641--9650}.
\newblock


\bibitem[\protect\citeauthoryear{Li, Chen, Xie, Yang, Yuan, Pu, and Zhuang}{Li
  et~al\mbox{.}}{2020b}]%
        {li2020free}
\bibfield{author}{\bibinfo{person}{Xianfeng Li}, \bibinfo{person}{Weijie Chen},
  \bibinfo{person}{Di Xie}, \bibinfo{person}{Shicai Yang},
  \bibinfo{person}{Peng Yuan}, \bibinfo{person}{Shiliang Pu}, {and}
  \bibinfo{person}{Yueting Zhuang}.} \bibinfo{year}{2020}\natexlab{b}.
\newblock \showarticletitle{A Free Lunch for Unsupervised Domain Adaptive
  Object Detection without Source Data}.
\newblock \bibinfo{journal}{\emph{arXiv preprint arXiv:2012.05400}}
  (\bibinfo{year}{2020}).
\newblock


\bibitem[\protect\citeauthoryear{Lian, Lv, Duan, and Gong}{Lian
  et~al\mbox{.}}{2019}]%
        {lian2019constructing}
\bibfield{author}{\bibinfo{person}{Qing Lian}, \bibinfo{person}{Fengmao Lv},
  \bibinfo{person}{Lixin Duan}, {and} \bibinfo{person}{Boqing Gong}.}
  \bibinfo{year}{2019}\natexlab{}.
\newblock \showarticletitle{Constructing self-motivated pyramid curriculums for
  cross-domain semantic segmentation: A non-adversarial approach}. In
  \bibinfo{booktitle}{\emph{Proceedings of the IEEE/CVF International
  Conference on Computer Vision}}. \bibinfo{pages}{6758--6767}.
\newblock


\bibitem[\protect\citeauthoryear{Liang, Hu, and Feng}{Liang
  et~al\mbox{.}}{2020}]%
        {liang2020we}
\bibfield{author}{\bibinfo{person}{Jian Liang}, \bibinfo{person}{Dapeng Hu},
  {and} \bibinfo{person}{Jiashi Feng}.} \bibinfo{year}{2020}\natexlab{}.
\newblock \showarticletitle{Do We Really Need to Access the Source Data? Source
  Hypothesis Transfer for Unsupervised Domain Adaptation}.
\newblock \bibinfo{journal}{\emph{arXiv preprint arXiv:2002.08546}}
  (\bibinfo{year}{2020}).
\newblock


\bibitem[\protect\citeauthoryear{Long, Cao, Wang, and Jordan}{Long
  et~al\mbox{.}}{2015}]%
        {long2015learning}
\bibfield{author}{\bibinfo{person}{Mingsheng Long}, \bibinfo{person}{Yue Cao},
  \bibinfo{person}{Jianmin Wang}, {and} \bibinfo{person}{Michael~I Jordan}.}
  \bibinfo{year}{2015}\natexlab{}.
\newblock \showarticletitle{Learning transferable features with deep adaptation
  networks}.
\newblock \bibinfo{journal}{\emph{arXiv preprint arXiv:1502.02791}}
  (\bibinfo{year}{2015}).
\newblock


\bibitem[\protect\citeauthoryear{Luo, Liu, Guan, Yu, and Yang}{Luo
  et~al\mbox{.}}{2019a}]%
        {luo2019significance}
\bibfield{author}{\bibinfo{person}{Yawei Luo}, \bibinfo{person}{Ping Liu},
  \bibinfo{person}{Tao Guan}, \bibinfo{person}{Junqing Yu}, {and}
  \bibinfo{person}{Yi Yang}.} \bibinfo{year}{2019}\natexlab{a}.
\newblock \showarticletitle{Significance-aware information bottleneck for
  domain adaptive semantic segmentation}. In
  \bibinfo{booktitle}{\emph{Proceedings of the IEEE/CVF International
  Conference on Computer Vision}}. \bibinfo{pages}{6778--6787}.
\newblock


\bibitem[\protect\citeauthoryear{Luo, Liu, Zheng, Guan, Yu, and Yang}{Luo
  et~al\mbox{.}}{2021}]%
        {luo2021category}
\bibfield{author}{\bibinfo{person}{Yawei Luo}, \bibinfo{person}{Ping Liu},
  \bibinfo{person}{Liang Zheng}, \bibinfo{person}{Tao Guan},
  \bibinfo{person}{Junqing Yu}, {and} \bibinfo{person}{Yi Yang}.}
  \bibinfo{year}{2021}\natexlab{}.
\newblock \showarticletitle{Category-Level Adversarial Adaptation for Semantic
  Segmentation using Purified Features}.
\newblock \bibinfo{journal}{\emph{IEEE Transactions on Pattern Analysis and
  Machine Intelligence}} (\bibinfo{year}{2021}).
\newblock


\bibitem[\protect\citeauthoryear{Luo, Zheng, Guan, Yu, and Yang}{Luo
  et~al\mbox{.}}{2019b}]%
        {luo2019taking}
\bibfield{author}{\bibinfo{person}{Yawei Luo}, \bibinfo{person}{Liang Zheng},
  \bibinfo{person}{Tao Guan}, \bibinfo{person}{Junqing Yu}, {and}
  \bibinfo{person}{Yi Yang}.} \bibinfo{year}{2019}\natexlab{b}.
\newblock \showarticletitle{Taking a closer look at domain shift:
  Category-level adversaries for semantics consistent domain adaptation}. In
  \bibinfo{booktitle}{\emph{Proceedings of the IEEE/CVF Conference on Computer
  Vision and Pattern Recognition}}. \bibinfo{pages}{2507--2516}.
\newblock


\bibitem[\protect\citeauthoryear{Mahmood, Chen, and Durr}{Mahmood
  et~al\mbox{.}}{2018}]%
        {mahmood2018unsupervised}
\bibfield{author}{\bibinfo{person}{Faisal Mahmood}, \bibinfo{person}{Richard
  Chen}, {and} \bibinfo{person}{Nicholas~J Durr}.}
  \bibinfo{year}{2018}\natexlab{}.
\newblock \showarticletitle{Unsupervised reverse domain adaptation for
  synthetic medical images via adversarial training}.
\newblock \bibinfo{journal}{\emph{IEEE transactions on medical imaging}}
  \bibinfo{volume}{37}, \bibinfo{number}{12} (\bibinfo{year}{2018}),
  \bibinfo{pages}{2572--2581}.
\newblock


\bibitem[\protect\citeauthoryear{Pan, Shin, Rameau, Lee, and Kweon}{Pan
  et~al\mbox{.}}{2020}]%
        {pan2020unsupervised}
\bibfield{author}{\bibinfo{person}{Fei Pan}, \bibinfo{person}{Inkyu Shin},
  \bibinfo{person}{Francois Rameau}, \bibinfo{person}{Seokju Lee}, {and}
  \bibinfo{person}{In~So Kweon}.} \bibinfo{year}{2020}\natexlab{}.
\newblock \showarticletitle{Unsupervised intra-domain adaptation for semantic
  segmentation through self-supervision}. In
  \bibinfo{booktitle}{\emph{Proceedings of the IEEE/CVF Conference on Computer
  Vision and Pattern Recognition}}. \bibinfo{pages}{3764--3773}.
\newblock


\bibitem[\protect\citeauthoryear{Richter, Vineet, Roth, and Koltun}{Richter
  et~al\mbox{.}}{2016}]%
        {richter2016playing}
\bibfield{author}{\bibinfo{person}{Stephan~R Richter}, \bibinfo{person}{Vibhav
  Vineet}, \bibinfo{person}{Stefan Roth}, {and} \bibinfo{person}{Vladlen
  Koltun}.} \bibinfo{year}{2016}\natexlab{}.
\newblock \showarticletitle{Playing for data: Ground truth from computer
  games}. In \bibinfo{booktitle}{\emph{European conference on computer
  vision}}. Springer, \bibinfo{pages}{102--118}.
\newblock


\bibitem[\protect\citeauthoryear{Ros, Sellart, Materzynska, Vazquez, and
  Lopez}{Ros et~al\mbox{.}}{2016}]%
        {ros2016synthia}
\bibfield{author}{\bibinfo{person}{German Ros}, \bibinfo{person}{Laura
  Sellart}, \bibinfo{person}{Joanna Materzynska}, \bibinfo{person}{David
  Vazquez}, {and} \bibinfo{person}{Antonio~M Lopez}.}
  \bibinfo{year}{2016}\natexlab{}.
\newblock \showarticletitle{The synthia dataset: A large collection of
  synthetic images for semantic segmentation of urban scenes}. In
  \bibinfo{booktitle}{\emph{Proceedings of the IEEE conference on computer
  vision and pattern recognition}}. \bibinfo{pages}{3234--3243}.
\newblock


\bibitem[\protect\citeauthoryear{Saito, Watanabe, Ushiku, and Harada}{Saito
  et~al\mbox{.}}{2018}]%
        {saito2018maximum}
\bibfield{author}{\bibinfo{person}{Kuniaki Saito}, \bibinfo{person}{Kohei
  Watanabe}, \bibinfo{person}{Yoshitaka Ushiku}, {and} \bibinfo{person}{Tatsuya
  Harada}.} \bibinfo{year}{2018}\natexlab{}.
\newblock \showarticletitle{Maximum classifier discrepancy for unsupervised
  domain adaptation}. In \bibinfo{booktitle}{\emph{Proceedings of the IEEE
  Conference on Computer Vision and Pattern Recognition}}.
  \bibinfo{pages}{3723--3732}.
\newblock


\bibitem[\protect\citeauthoryear{Sakaridis, Dai, and Gool}{Sakaridis
  et~al\mbox{.}}{2019}]%
        {sakaridis2019guided}
\bibfield{author}{\bibinfo{person}{Christos Sakaridis},
  \bibinfo{person}{Dengxin Dai}, {and} \bibinfo{person}{Luc~Van Gool}.}
  \bibinfo{year}{2019}\natexlab{}.
\newblock \showarticletitle{Guided curriculum model adaptation and
  uncertainty-aware evaluation for semantic nighttime image segmentation}. In
  \bibinfo{booktitle}{\emph{Proceedings of the IEEE/CVF International
  Conference on Computer Vision}}. \bibinfo{pages}{7374--7383}.
\newblock


\bibitem[\protect\citeauthoryear{Springenberg}{Springenberg}{2015}]%
        {springenberg2015unsupervised}
\bibfield{author}{\bibinfo{person}{Jost~Tobias Springenberg}.}
  \bibinfo{year}{2015}\natexlab{}.
\newblock \showarticletitle{Unsupervised and semi-supervised learning with
  categorical generative adversarial networks}.
\newblock \bibinfo{journal}{\emph{arXiv preprint arXiv:1511.06390}}
  (\bibinfo{year}{2015}).
\newblock


\bibitem[\protect\citeauthoryear{Sun and Saenko}{Sun and Saenko}{2016}]%
        {sun2016deep}
\bibfield{author}{\bibinfo{person}{Baochen Sun} {and} \bibinfo{person}{Kate
  Saenko}.} \bibinfo{year}{2016}\natexlab{}.
\newblock \showarticletitle{Deep coral: Correlation alignment for deep domain
  adaptation}. In \bibinfo{booktitle}{\emph{European conference on computer
  vision}}. Springer, \bibinfo{pages}{443--450}.
\newblock


\bibitem[\protect\citeauthoryear{Tsai, Hung, Schulter, Sohn, Yang, and
  Chandraker}{Tsai et~al\mbox{.}}{2018}]%
        {tsai2018learning}
\bibfield{author}{\bibinfo{person}{Yi-Hsuan Tsai}, \bibinfo{person}{Wei-Chih
  Hung}, \bibinfo{person}{Samuel Schulter}, \bibinfo{person}{Kihyuk Sohn},
  \bibinfo{person}{Ming-Hsuan Yang}, {and} \bibinfo{person}{Manmohan
  Chandraker}.} \bibinfo{year}{2018}\natexlab{}.
\newblock \showarticletitle{Learning to adapt structured output space for
  semantic segmentation}. In \bibinfo{booktitle}{\emph{Proceedings of the IEEE
  conference on computer vision and pattern recognition}}.
  \bibinfo{pages}{7472--7481}.
\newblock


\bibitem[\protect\citeauthoryear{Tsai, Sohn, Schulter, and Chandraker}{Tsai
  et~al\mbox{.}}{2019}]%
        {tsai2019domain}
\bibfield{author}{\bibinfo{person}{Yi-Hsuan Tsai}, \bibinfo{person}{Kihyuk
  Sohn}, \bibinfo{person}{Samuel Schulter}, {and} \bibinfo{person}{Manmohan
  Chandraker}.} \bibinfo{year}{2019}\natexlab{}.
\newblock \showarticletitle{Domain adaptation for structured output via
  discriminative patch representations}. In
  \bibinfo{booktitle}{\emph{Proceedings of the IEEE/CVF International
  Conference on Computer Vision}}. \bibinfo{pages}{1456--1465}.
\newblock


\bibitem[\protect\citeauthoryear{Vu, Jain, Bucher, Cord, and P{\'e}rez}{Vu
  et~al\mbox{.}}{2019}]%
        {vu2019advent}
\bibfield{author}{\bibinfo{person}{Tuan-Hung Vu}, \bibinfo{person}{Himalaya
  Jain}, \bibinfo{person}{Maxime Bucher}, \bibinfo{person}{Matthieu Cord},
  {and} \bibinfo{person}{Patrick P{\'e}rez}.} \bibinfo{year}{2019}\natexlab{}.
\newblock \showarticletitle{Advent: Adversarial entropy minimization for domain
  adaptation in semantic segmentation}. In
  \bibinfo{booktitle}{\emph{Proceedings of the IEEE/CVF Conference on Computer
  Vision and Pattern Recognition}}. \bibinfo{pages}{2517--2526}.
\newblock


\bibitem[\protect\citeauthoryear{Wang, Yang, Xu, Hanjalic, and Shen}{Wang
  et~al\mbox{.}}{2017}]%
        {wang2017adversarial}
\bibfield{author}{\bibinfo{person}{Bokun Wang}, \bibinfo{person}{Yang Yang},
  \bibinfo{person}{Xing Xu}, \bibinfo{person}{Alan Hanjalic}, {and}
  \bibinfo{person}{Heng~Tao Shen}.} \bibinfo{year}{2017}\natexlab{}.
\newblock \showarticletitle{Adversarial cross-modal retrieval}. In
  \bibinfo{booktitle}{\emph{Proceedings of the 25th ACM international
  conference on Multimedia}}. \bibinfo{pages}{154--162}.
\newblock


\bibitem[\protect\citeauthoryear{Wang, Yu, Wei, Feris, Xiong, Hwu, Huang, and
  Shi}{Wang et~al\mbox{.}}{2020}]%
        {wang2020differential}
\bibfield{author}{\bibinfo{person}{Zhonghao Wang}, \bibinfo{person}{Mo Yu},
  \bibinfo{person}{Yunchao Wei}, \bibinfo{person}{Rogerio Feris},
  \bibinfo{person}{Jinjun Xiong}, \bibinfo{person}{Wen-mei Hwu},
  \bibinfo{person}{Thomas~S Huang}, {and} \bibinfo{person}{Honghui Shi}.}
  \bibinfo{year}{2020}\natexlab{}.
\newblock \showarticletitle{Differential treatment for stuff and things: A
  simple unsupervised domain adaptation method for semantic segmentation}. In
  \bibinfo{booktitle}{\emph{Proceedings of the IEEE/CVF Conference on Computer
  Vision and Pattern Recognition}}. \bibinfo{pages}{12635--12644}.
\newblock


\bibitem[\protect\citeauthoryear{Yang and Soatto}{Yang and Soatto}{2020}]%
        {yang2020fda}
\bibfield{author}{\bibinfo{person}{Yanchao Yang} {and} \bibinfo{person}{Stefano
  Soatto}.} \bibinfo{year}{2020}\natexlab{}.
\newblock \showarticletitle{Fda: Fourier domain adaptation for semantic
  segmentation}. In \bibinfo{booktitle}{\emph{Proceedings of the IEEE/CVF
  Conference on Computer Vision and Pattern Recognition}}.
  \bibinfo{pages}{4085--4095}.
\newblock


\bibitem[\protect\citeauthoryear{You, Su, Li, Zhu, Lu, and Yang}{You
  et~al\mbox{.}}{2021}]%
        {you2021learning}
\bibfield{author}{\bibinfo{person}{Fuming You}, \bibinfo{person}{Hongzu Su},
  \bibinfo{person}{Jingjing Li}, \bibinfo{person}{Lei Zhu}, \bibinfo{person}{Ke
  Lu}, {and} \bibinfo{person}{Yang Yang}.} \bibinfo{year}{2021}\natexlab{}.
\newblock \showarticletitle{Learning a Weighted Classifier for conditional
  domain adaptation}.
\newblock \bibinfo{journal}{\emph{Knowledge-Based Systems}}
  (\bibinfo{year}{2021}), \bibinfo{pages}{106774}.
\newblock


\bibitem[\protect\citeauthoryear{Yu, Zhang, Dong, Hu, Dong, and Zhang}{Yu
  et~al\mbox{.}}{2021}]%
        {yu2021dast}
\bibfield{author}{\bibinfo{person}{Fei Yu}, \bibinfo{person}{Mo Zhang},
  \bibinfo{person}{Hexin Dong}, \bibinfo{person}{Sheng Hu},
  \bibinfo{person}{Bin Dong}, {and} \bibinfo{person}{Li Zhang}.}
  \bibinfo{year}{2021}\natexlab{}.
\newblock \showarticletitle{DAST: Unsupervised Domain Adaptation in Semantic
  Segmentation Based on Discriminator Attention and Self-Training}. In
  \bibinfo{booktitle}{\emph{Proceedings of the AAAI Conference on Artificial
  Intelligence}}, Vol.~\bibinfo{volume}{35}. \bibinfo{pages}{10754--10762}.
\newblock


\bibitem[\protect\citeauthoryear{Zhang, Zhang, Liu, and Tao}{Zhang
  et~al\mbox{.}}{2019}]%
        {zhang2019category}
\bibfield{author}{\bibinfo{person}{Qiming Zhang}, \bibinfo{person}{Jing Zhang},
  \bibinfo{person}{Wei Liu}, {and} \bibinfo{person}{Dacheng Tao}.}
  \bibinfo{year}{2019}\natexlab{}.
\newblock \showarticletitle{Category anchor-guided unsupervised domain
  adaptation for semantic segmentation}.
\newblock \bibinfo{journal}{\emph{arXiv preprint arXiv:1910.13049}}
  (\bibinfo{year}{2019}).
\newblock


\bibitem[\protect\citeauthoryear{Zhang, Qiu, Yao, Liu, and Mei}{Zhang
  et~al\mbox{.}}{2018}]%
        {zhang2018fully}
\bibfield{author}{\bibinfo{person}{Yiheng Zhang}, \bibinfo{person}{Zhaofan
  Qiu}, \bibinfo{person}{Ting Yao}, \bibinfo{person}{Dong Liu}, {and}
  \bibinfo{person}{Tao Mei}.} \bibinfo{year}{2018}\natexlab{}.
\newblock \showarticletitle{Fully convolutional adaptation networks for
  semantic segmentation}. In \bibinfo{booktitle}{\emph{Proceedings of the IEEE
  Conference on Computer Vision and Pattern Recognition}}.
  \bibinfo{pages}{6810--6818}.
\newblock


\bibitem[\protect\citeauthoryear{Zhao, Des~Combes, Zhang, and Gordon}{Zhao
  et~al\mbox{.}}{2019}]%
        {zhao2019learning}
\bibfield{author}{\bibinfo{person}{Han Zhao}, \bibinfo{person}{Remi~Tachet
  Des~Combes}, \bibinfo{person}{Kun Zhang}, {and} \bibinfo{person}{Geoffrey
  Gordon}.} \bibinfo{year}{2019}\natexlab{}.
\newblock \showarticletitle{On Learning Invariant Representations for Domain
  Adaptation}. In \bibinfo{booktitle}{\emph{International Conference on Machine
  Learning}}. \bibinfo{pages}{7523--7532}.
\newblock


\bibitem[\protect\citeauthoryear{Zheng and Yang}{Zheng and Yang}{2021}]%
        {zheng2021rectifying}
\bibfield{author}{\bibinfo{person}{Zhedong Zheng} {and} \bibinfo{person}{Yi
  Yang}.} \bibinfo{year}{2021}\natexlab{}.
\newblock \showarticletitle{Rectifying pseudo label learning via uncertainty
  estimation for domain adaptive semantic segmentation}.
\newblock \bibinfo{journal}{\emph{International Journal of Computer Vision}}
  (\bibinfo{year}{2021}), \bibinfo{pages}{1--15}.
\newblock


\bibitem[\protect\citeauthoryear{Zou, Yu, Kumar, and Wang}{Zou
  et~al\mbox{.}}{2018}]%
        {zou2018unsupervised}
\bibfield{author}{\bibinfo{person}{Yang Zou}, \bibinfo{person}{Zhiding Yu},
  \bibinfo{person}{BVK Kumar}, {and} \bibinfo{person}{Jinsong Wang}.}
  \bibinfo{year}{2018}\natexlab{}.
\newblock \showarticletitle{Unsupervised domain adaptation for semantic
  segmentation via class-balanced self-training}. In
  \bibinfo{booktitle}{\emph{Proceedings of the European conference on computer
  vision (ECCV)}}. \bibinfo{pages}{289--305}.
\newblock


\bibitem[\protect\citeauthoryear{Zou, Yu, Liu, Kumar, and Wang}{Zou
  et~al\mbox{.}}{2019}]%
        {zou2019confidence}
\bibfield{author}{\bibinfo{person}{Yang Zou}, \bibinfo{person}{Zhiding Yu},
  \bibinfo{person}{Xiaofeng Liu}, \bibinfo{person}{BVK Kumar}, {and}
  \bibinfo{person}{Jinsong Wang}.} \bibinfo{year}{2019}\natexlab{}.
\newblock \showarticletitle{Confidence regularized self-training}. In
  \bibinfo{booktitle}{\emph{Proceedings of the IEEE/CVF International
  Conference on Computer Vision}}. \bibinfo{pages}{5982--5991}.
\newblock


\end{thebibliography}

			\end{document}